\documentclass[11pt,twocolumn,letterpaper]{article}

\usepackage{iccv}
\usepackage{times}
\usepackage{epsfig}
\usepackage{graphicx}
\usepackage{amsmath}
\usepackage{amssymb}
\usepackage{siunitx}
\usepackage{gensymb}

\usepackage{tabularx}
\usepackage{multirow}
\usepackage{booktabs}  
\usepackage{colortbl}  
\newcolumntype{C}{>{\centering\arraybackslash}X}

\usepackage{xcolor,colortbl}

\usepackage[pagebackref=true,breaklinks=true,letterpaper=true,colorlinks,bookmarks=false]{hyperref}

\iccvfinalcopy

\definecolor{LightGray}{rgb}{0.95,0.95,0.95}
\usepackage{color, colortbl}
\usepackage{amsmath}
\usepackage{dsfont}
\usepackage{subcaption}
\usepackage{afterpage} 
\usepackage{hyperref}
\usepackage{placeins} 
\usepackage{graphicx}
\usepackage{makecell}
\usepackage{booktabs}

\DeclareMathOperator*{\argmax}{argmax} %

\newcommand{\pic}[2]{
	\begin{figure}[htb] 
		\centering
		\includegraphics[width=\linewidth]{#1}
		\caption{#2}
		\label{fig:#1}
	\end{figure}
}

\newcommand{\picLarge}[2]{
	\begin{figure*}[htb] 
		\centering
		\includegraphics[width=\textwidth]{#1}
		\caption{#2}
		\label{fig:#1}
	\end{figure*}
}

\newcommand{\picFour}[6]{
	\begin{figure*}[htb]
		\begin{subfigure}[c]{0.246\textwidth}        
			\includegraphics[width=\textwidth]{#1-0}
			\subcaption{#2}
			\label{fig:#1-0}        
		\end{subfigure}
		\begin{subfigure}[c]{0.246\textwidth}        
			\includegraphics[width=\textwidth]{#1-1}
			\subcaption{#3}  
			\label{fig:#1-1}       
		\end{subfigure}
		\begin{subfigure}[c]{0.246\textwidth}        
			\includegraphics[width=\textwidth]{#1-2}
			\subcaption{#4}  
			\label{fig:#1-2}       
		\end{subfigure}
		\begin{subfigure}[c]{0.246\textwidth}        
			\includegraphics[width=\textwidth]{#1-3}
			\subcaption{#5}
			\label{fig:#1-3}         
		\end{subfigure}    
		\caption{#6}
		\label{fig:#1}
	\end{figure*}
}

\newcommand{\picEight}[9]{
	\begin{figure*}[htb]
		\begin{subfigure}[c]{0.246\textwidth}        
			\includegraphics[width=\textwidth]{#1-0}
			\subcaption{#2}        
		\end{subfigure}
		\begin{subfigure}[c]{0.246\textwidth}        
			\includegraphics[width=\textwidth]{#1-1}
			\subcaption{#3}        
		\end{subfigure}
		\begin{subfigure}[c]{0.246\textwidth}        
			\includegraphics[width=\textwidth]{#1-2}
			\subcaption{#4}        
		\end{subfigure}
		\begin{subfigure}[c]{0.246\textwidth}        
			\includegraphics[width=\textwidth]{#1-3}
			\subcaption{#5}        
		\end{subfigure}    
		\begin{subfigure}[c]{0.246\textwidth}        
			\includegraphics[width=\textwidth]{#1-4}
			\subcaption{#6}        
		\end{subfigure}
		\begin{subfigure}[c]{0.246\textwidth}        
			\includegraphics[width=\textwidth]{#1-5}
			\subcaption{#7}        
		\end{subfigure}
		\begin{subfigure}[c]{0.246\textwidth}        
			\includegraphics[width=\textwidth]{#1-6}
			\subcaption{#8}        
		\end{subfigure}
		\begin{subfigure}[c]{0.246\textwidth}        
			\includegraphics[width=\textwidth]{#1-7}
			\subcaption{Contrastive Learning}        
		\end{subfigure}    
		\caption{#9}
		\label{fig:#1}
	\end{figure*}
}

\newcommand{\picThree}[5]{
	\begin{figure*}[htb]
		\begin{subfigure}[c]{0.32\textwidth}        
			\includegraphics[width=\textwidth]{#1-0}
			\subcaption{#2}        
		\end{subfigure}
		\begin{subfigure}[c]{0.32\textwidth}        
			\includegraphics[width=\textwidth]{#1-1}
			\subcaption{#3}        
		\end{subfigure}
		\begin{subfigure}[c]{0.32\textwidth}        
			\includegraphics[width=\textwidth]{#1-2}
			\subcaption{#4}        
		\end{subfigure}
		\caption{#5}
		\label{fig:#1}
	\end{figure*}
}

\begin{document}

	\title{A Survey on Semi-, Self- and Unsupervised Learning in Image Classification}

\author{Lars Schmarje\thanks{Corresponding author}, Monty Santarossa, Simon-Martin Schr\"oder, Reinhard Koch\\
	Multimedia Information Processing Group, Kiel University, Germany \\
	{\tt\small \{las,msa,sms,rk\}@informatik.uni-kiel.de}
}

\maketitle

\begin{abstract}
	While deep learning strategies achieve outstanding results in computer vision tasks, one issue remains:
	The current strategies rely heavily on a huge amount of labeled data.
	In many real-world problems, it is not feasible to create such an amount of labeled training data.
	Therefore, it is common to incorporate unlabeled data into the training process to reach equal results with fewer labels.
	Due to a lot of concurrent research, it is difficult to keep track of recent developments.
	In this survey, we provide an overview of often used ideas and methods in image classification with fewer labels.
	We compare 34 methods in detail based on their performance and their commonly used ideas rather than a fine-grained taxonomy.
	In our analysis, we identify three major trends that lead to future research opportunities.
	1. State-of-the-art methods are scaleable to real-world applications in theory but issues like class imbalance, robustness, or fuzzy labels are not considered. 
	2. The degree of supervision which is needed to achieve comparable results to the usage of all labels is decreasing and therefore methods need to be extended to settings with a variable number of classes.
	3. All methods share some common ideas but we identify clusters of methods that do not share many ideas.
	We show that combining ideas from different clusters can lead to better performance.
\end{abstract}

\vspace{-0.5cm}
\section{Introduction}
\label{sec:intro}
Deep learning strategies achieve outstanding successes in computer vision tasks.
They reach the best performance in a diverse range of tasks such as image classification \cite{imagenet,resnet,tailception}, object detection \cite{yolov3,parcel_tracking} or semantic segmentation \cite{fcn,maskrcnn}.
\pic{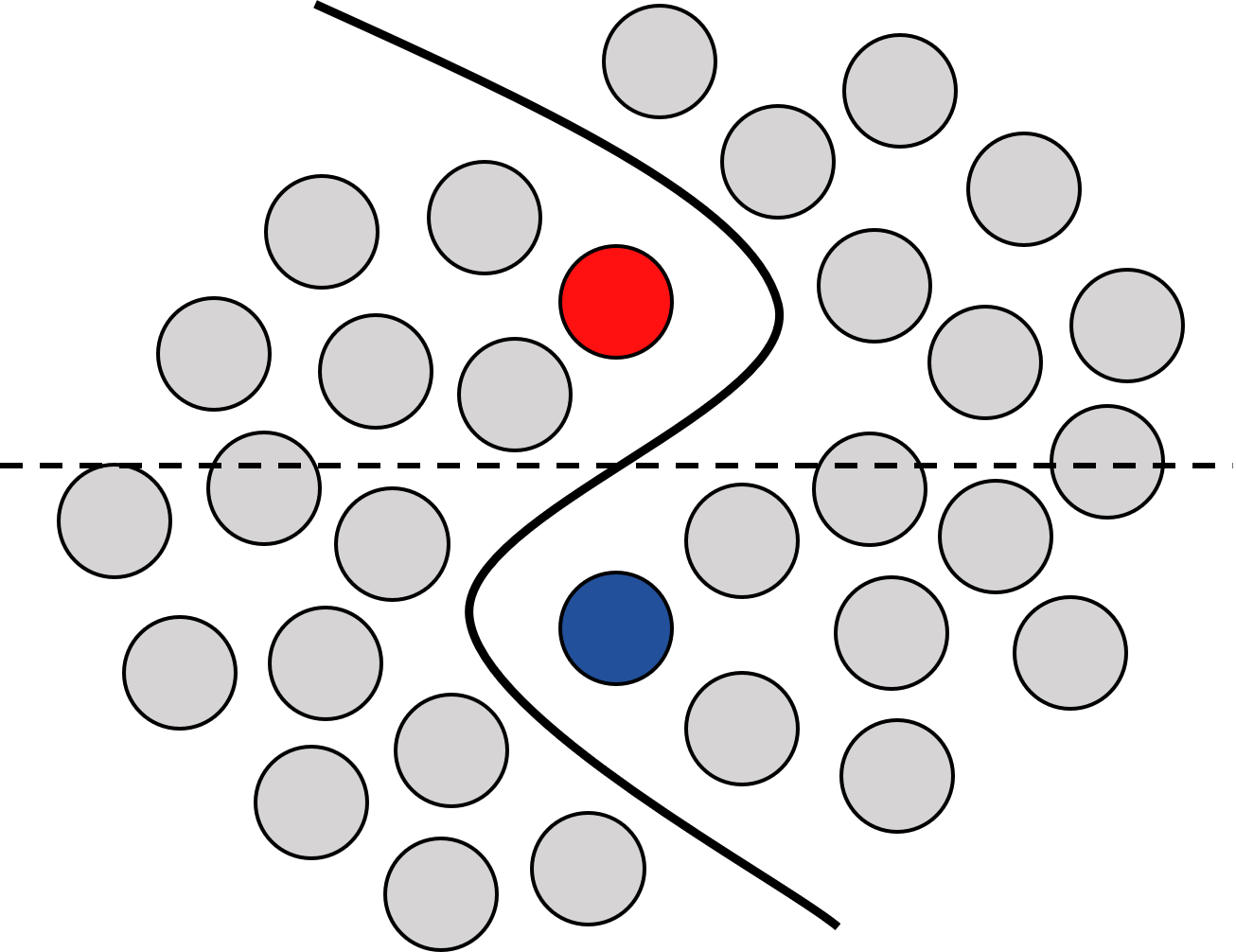}{
	This image illustrates and simplifies the benefit of using unlabeled data during deep learning training.
	The red and dark blue circles represent labeled data points of different classes.
	The light grey circles represent unlabeled data points.
	If we have only a small number of labeled samples available we can only make assumptions (dotted line) over the underlying true distribution (solid line).
	This true distribution can only be determined if we also consider the unlabeled data points and clarify the decision boundary.
}

The quality of a deep neural network is strongly influenced by the number of labeled/supervised images \cite{book-pattern}.
ImageNet \cite{imagenet} is a huge labeled dataset with over one million images which allows the training of networks with impressive performance.
Recent research shows that even larger datasets than ImageNet can improve these results \cite{limits-supervision}.
However, in many real-world applications it is not possible to create labeled datasets with millions of images.
A common strategy for dealing with this problem is transfer learning.
This strategy improves results even on small and specialized datasets like medical imaging \cite{schmarje2019}.
This might be a practical workaround for some applications but the fundamental issue remains:
Unlike humans, supervised learning needs enormous amounts of labeled data.

For a given problem we often have access to a large dataset of unlabeled data.
How this unsupervised data could be used for neural networks has been of research interest for many years \cite{book-unsupervised}.
Xie \etal were among the first in 2016 to investigate unsupervised deep learning image clustering strategies to leverage this data \cite{dec}.
Since then, the usage of unlabeled data has been researched in numerous ways and has created research fields like unsupervised, semi-supervised, self-supervised, weakly-supervised, or metric learning \cite{metric-survey}.
Generally speaking, unsupervised learning uses no labeled data, semi-supervised learning uses unlabeled and labeled while self-supervised learning generates labeled data on its own.
Other research directions are even more different because weakly-supervised learning uses only partial information about the label and metric learning aims at learning a good distance metric.
The idea that unifies these approaches is that using unlabeled data is beneficial during the training process (see \autoref{fig:concept-semi-supervised-learning} for an illustration).
It either makes the training with fewer labels more robust or in some rare cases even surpasses the supervised cases \cite{iic}.

Due to this benefit, many researchers and companies work in the field of semi-, self-, and unsupervised learning.
The main goal is to close the gap between semi-supervised and supervised learning or even surpass these results.
Considering presented methods like \cite{S4L,uda} we believe that research is at the breaking point of achieving this goal.
Hence, there is a lot of research ongoing in this field.
This survey provides an overview to keep track of the major and recent developments in semi-, self-, and unsupervised learning.

Most investigated research topics share a variety of common ideas while differing in goal, application contexts, and implementation details.
This survey gives an overview of this wide range of research topics.
The focus of this survey is on describing the similarities and differences between the methods.

Whereas we look at a broad range of learning strategies, we compare these methods only based on the image classification task.
The addressed audience of this survey consists of deep learning researchers or interested people with comparable preliminary knowledge who want to keep track of recent developments in the field of \mbox{semi-,} self- and unsupervised learning.

\subsection{Related Work}

\picLarge{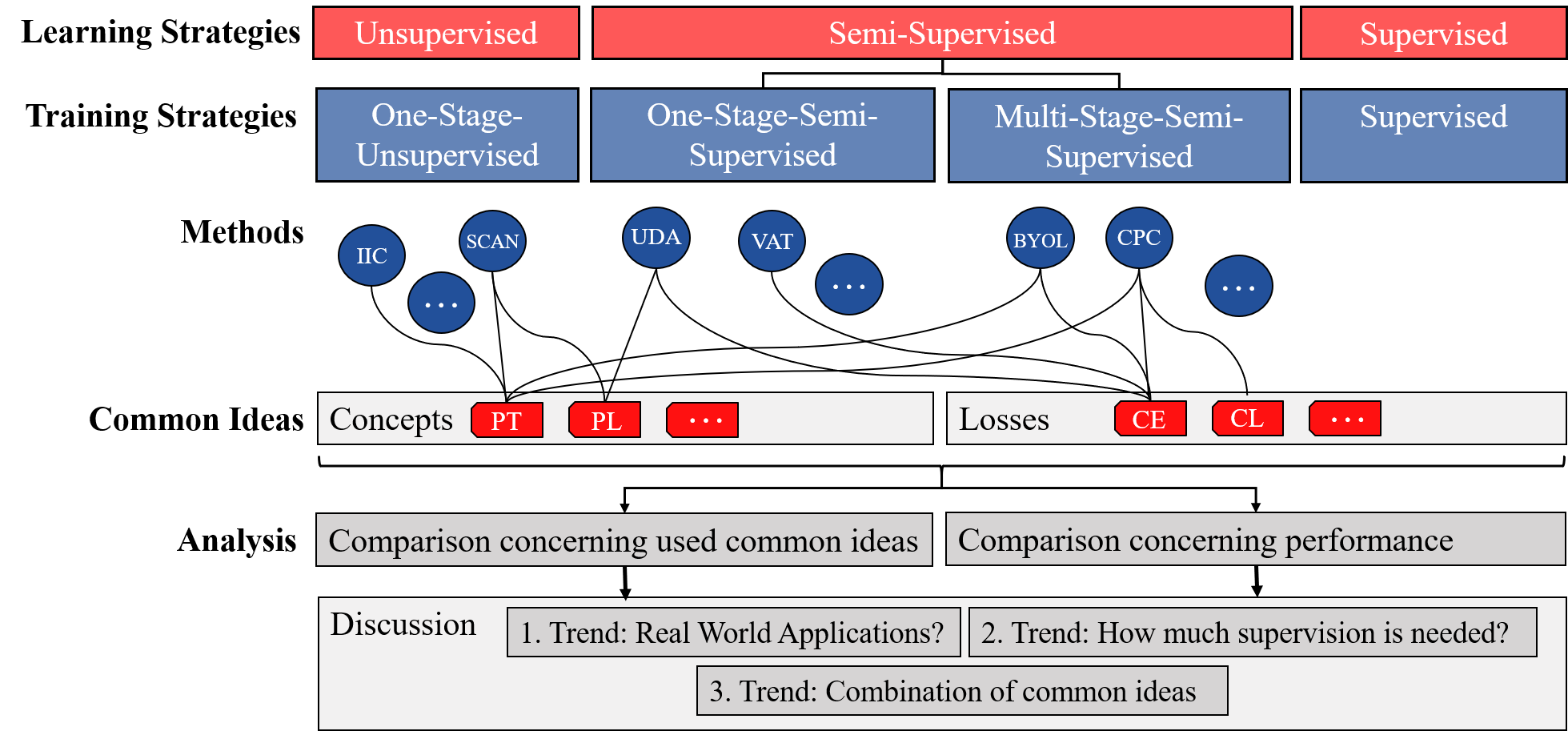}{Overview of the structure of this survey --
	The learning strategies unsupervised, semi-supervised and supervised are commonly used in the literature. 
	Because semi-supervised learning is incorporating many methods we defined training strategies which subdivides semi-supervised learning.
	For details about the training and learning strategies (including self-supervised learning) see \autoref{subsec:strategies}.
	Each method belongs to one training strategy and uses several common ideas. 
	A common idea can be a concept such as a pretext task or a loss such as cross-entropy.
	The definition of methods and common ideas is given in \autoref{sec:pre}.
	Details about the common ideas are defined in \autoref{subsec:ideas}.
	All methods in this survey are shortly described and categorized in \autoref{sec:methods}.
	The methods are compared with each other based on this information concerning their used common ideas and their performance in \autoref{subsec:comparison}.
	The results of the comparisons and three resulting trends are discussed in \autoref{subsec:discus}.
}

In this subsection, we give a quick overview of previous works and reference topics we will not address further to maintain the focus of this survey.

The research of semi- and unsupervised techniques in computer vision has a long history.
A variety of research, surveys, and books has been published on this topic \cite{semi-supervised-learning,old-clustering,survey-self, survey-semi, unsupervised-feature}.
Unsupervised cluster algorithms were researched before the breakthrough of deep learning and are still widely used \cite{k-means}.
There are already extensive surveys that describe unsupervised and semi-supervised strategies without deep learning \cite{old-clustering,old-semi-supervised}.
We will focus only on techniques including deep neural networks.

Many newer surveys focus only on self-, semi- or unsupervised learning \cite{deep-survey,survey-self,survey-semi}.
Min \etal wrote an overview of unsupervised deep learning strategies \cite{deep-survey}.
They presented the beginning in this field of research from a network architecture perspective.
The authors looked at a broad range of architectures.
We focus on only one architecture which Min \etal refer to as  "Clustering deep neural network (CDNN)-based deep clustering" \cite{deep-survey}. 
Even though the work was published in 2018, it already misses the recent and major developments in deep learning of the last years.
We look at these more recent developments and show the connections to other research fields that Min \etal did not include.

Van Engelen and Hoos give a broad overview of general and recent semi-supervised methods \cite{survey-semi}.
They cover some recent developments but deep learning strategies such as \cite{simclr,fixmatch,foc,iic,byol} are not covered.
Furthermore, the authors do not explicitly compare the presented methods based on their structure or performance. 

Jing and Tian concentrated their survey on recent developments in self-supervised learning \cite{survey-self}.
Like us, the authors provide a performance comparison and a taxonomy.
Their taxonomy distinguishes between different kinds of pretext tasks.
We look at pretext tasks as one common idea and compare the methods based on these underlying ideas.
Jing and Tian look at different tasks apart from classification but do not include semi- and unsupervised methods without a pretext task.

Qi and Luo are one of the few who look at self-, semi- and unsupervised learning in one survey \cite{survey-semi-unsuper}. 
However, they look at the different learning strategies separately and give comparisons only inside the respective learning strategy.
We show that bridging these gaps leads to new insights, improved performance, and future research approaches.

Some surveys focus not on the general overviews about semi-, self-, and unsupervised learning but special details.
In their survey, Cheplygina \etal present a variety of methods in the context of medical image analysis \cite{medicine-survey}.
They include deep learning and older machine learning approaches but look at different strategies from a medical perspective. 
Mey and Loog focused on the underlying theoretical assumptions in semi-supervised learning \cite{semi-theory}.
We keep our survey limited to general image classification tasks and focus on their practical application.

In this survey, we will focus on deep learning approaches for image classification.
We will investigate the different learning strategies with a spotlight on loss functions.
We concentrate on recent methods because older one are already adequately addressed in previous literature \cite{semi-supervised-learning,old-clustering,survey-self, survey-semi, unsupervised-feature}.
Keeping the above-mentioned limitations in mind, the topic of self-, semi-, and unsupervised learning still includes a broad range of research fields.
We have to exclude some related topics from this survey to keep the focus of this work for example because other research have a different aim or are evaluated on different datasets.
Therefore, topics like metric learning \cite{metric-survey} and meta learning such as \cite{metalearning} will be excluded.
More specific networks like general adversarial networks \cite{gan} and graph networks such as \cite{graphnetwork} will be excluded.  
Also, other applications like pose estimation \cite{pose} and segmentation \cite{multi-segmentation} or other image sources like videos or sketches \cite{hand-sketch} are excluded.
Topics like few-shot or zero-shot learning methods such as \cite{zeroshot} are excluded in this survey.
However, we will see in \autoref{subsec:discus} that topics like few-shot learning and semi-supervised can learn from each other in the future like in \cite{transmatch}.

\subsection{Outline}

The rest of the paper is structured in the following way.
We define and explain the terms which are used in this survey such as method, training strategy and common idea in \autoref{sec:pre}.
A visual representation of the terms and their dependencies can be seen before the analysis part in \autoref{fig:overview}. 
All methods are presented with a short description, their training strategy and common idea in \autoref{sec:methods}.
In \autoref{sec:compare}, we compare the methods based their used ideas and their performance across four common image classification datasets.
This section also includes a description of the datasets and evaluation metrics.
Finally, we discuss the results of the comparisons in \autoref{subsec:discus} and identify three trends and research opportunities.
In \autoref{fig:overview}, a complete overview of the structure of this survey can be seen.
\\
\picFour{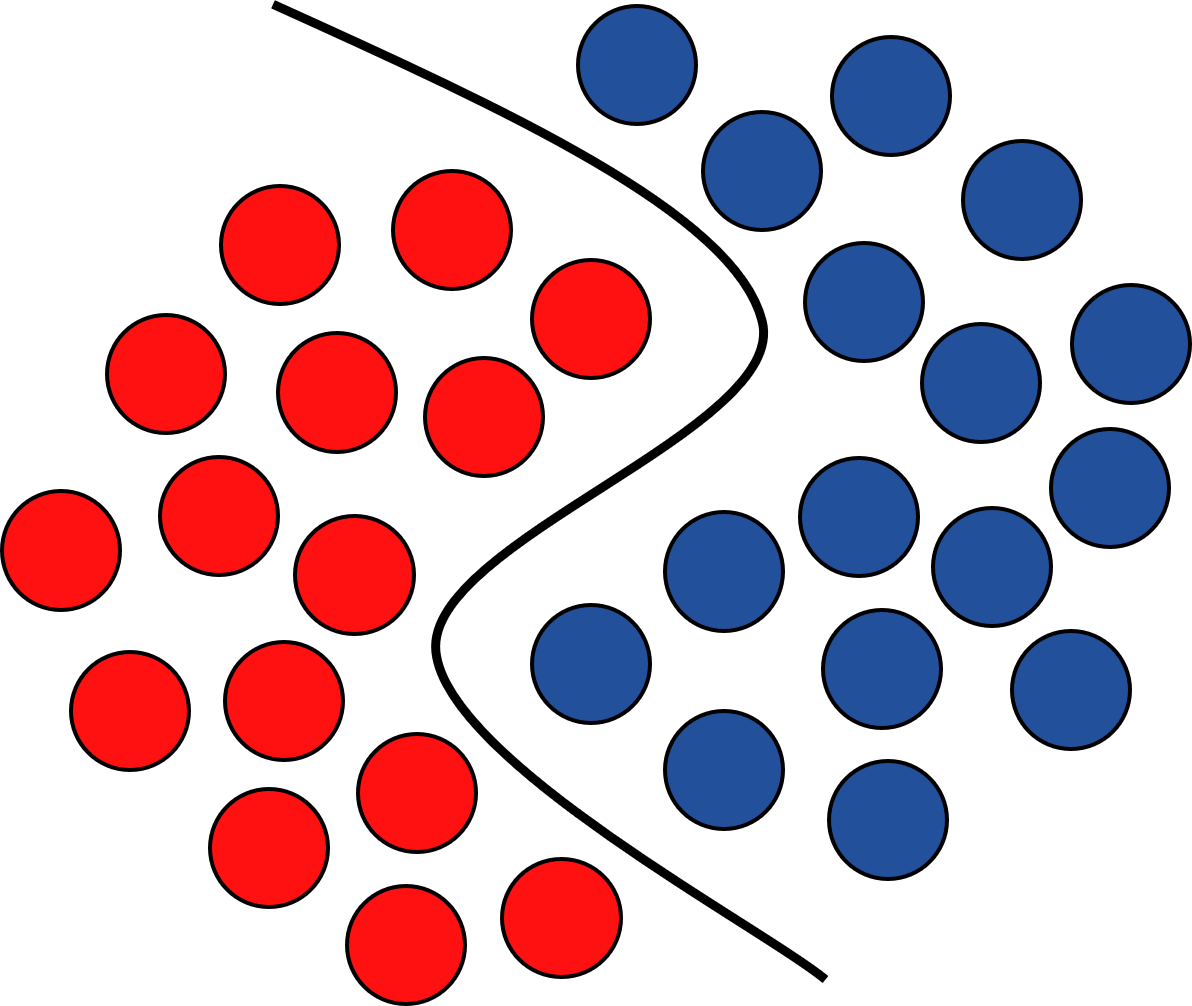}{\footnotesize Supervised}{\footnotesize One-Stage-Semi-Supervised}{\footnotesize One-Stage-Unsupervised}{\footnotesize Multi-Stage-Semi-Supervised}{Illustrations of supervised learning (a) and the three presented reduced training strategies (b-d) -
	The red and dark blue circles represent labeled data points of different classes.
	The light grey circles represent unlabeled data points. 
	The black lines define the underlying decision boundaries between the classes.
	The striped circles represent data points that do not use the label information in the first stage and can access this information in a second stage. 
	For more details on stages and the different learning strategies see \autoref{subsec:strategies}. }

\section{Underlying Concepts}
\label{sec:pre}
Throughout this survey, we use the terms training strategy, common idea, and method in a specific meaning.
The \emph{training strategy} is the general type/approach for using the unsupervised data during training.
The training strategies are similar to the terms semi-supervised, self-supervised, or unsupervised learning but provide a definition for corner cases that the other terms do not.
We will explain the differences and similarities in detail in \autoref{subsec:strategies}.
The papers we discuss in detail in this survey propose different elements like an algorithm, a general idea, or an extension of previous work.
To be consistent in this survey, we call the main algorithm, idea, or extension in each paper a \emph{method}.
All methods are briefly described in \autoref{sec:methods}.
A method follows a training strategy and is based on several \emph{common ideas}.
We use the term common idea, or in short idea, for concepts and approaches that are shared between different methods.
We roughly sort the methods based on their training strategy but compare them in detail based on the used common ideas.
See \autoref{subsec:ideas} for further information about common ideas. 

In the rest of this chapter,  we will use a shared definition for the following variables.
For an arbitrary set of images $X$ we define $X_l$ and  $X_u$ with  $X = {X_l \dot{\cup} X_u}$  as the labeled and unlabeled images, respectively.
For an image $x \in X_l$ the corresponding label is defined as $z_x \in Z$.
An image $x \in X_u$ has no label otherwise it would belong to $X_l$. 
For the distinction between $X_u$ and $X_l$, only the usage of the label information during training is important.
For example, an image $x \in X$ might have a label that can be used during evaluation but as long as the label is not used during training we define $x \in X_u$.
The learning strategy $LS_{X}$ for a dataset $X$ is either unsupervised ($X = X_u$), supervised ($X = X_l$) or semi-supervised ($X_u \cap X_l \neq \emptyset$).
During different phases of the training, different image datasets $X_1, X_2, \dots X_n$ with $n \in \mathbb{N}$ could be used. 
Two consecutive datasets $X_i$ and $X_{i+1}$ with $i \leq n$ and $i \in \mathbb{N}$ are different as long as different images ($X_i \neq X_{i+1}$) or different labels ($X_{L_i} \neq X_{L_{i+1}}$)  are used.
The learning strategy $LS_i$ up to the dataset $X_i$ during the training is calculated based on $X_u = \cup_{j=1}^{i} X_{u_{j}}$ and $X_l = \cup_{j=1}^{i} X_{l_{j}}$.
Consecutive phases of the training are grouped into \emph{stages}.
The stage changes during consecutive datasets $X_i$ and $X_{i+1}$ iff the learning strategy is different ($LS_{X_i} \neq LS_{X_{i+1}}$) and the overall learning strategy changes ($LS_i \neq LS_{i+1}$).
Due to this definition, only two stages can occur during training and the seven possible combinations are visualized in \autoref{fig:stages}.
For more details see \autoref{subsec:strategies}.
Let $C$ be the number of classes for the labels $Z$.
For a given neural network $f$ and input $x \in X$ the output of the neural network is $f(x)$.
For the below-defined formulations, $f$ is an arbitrary network with arbitrary weights and parameters.

\subsection{Training strategies}
\label{subsec:strategies}
Terms like semi-supervised, self-supervised, and unsupervised learning are often used in literature but have overlapping definitions for certain methods.
We will summarize the general understanding and definition of these terms and highlight borderline cases that are difficult to classify.
Due to these borderline cases, we will define a new taxonomy based on the stages during training for a precise distinction of the methods.
In \autoref{subsec:comparison}, we will see that this taxonomy leads to a clear clustering of the methods regarding the common ideas which further justifies this taxonomy.
A visual comparison between the learning-strategies semi-supervised and unsupervised learning and the training strategies can be found in \autoref{fig:stages}.

Unsupervised learning describes the training without any labels. 
However, the goal can be a clustering (e.g. \cite{iic,foc}) or good representation (e.g. \cite{simclr,self-rotation}) of the data.
Some methods combine several unsupervised steps to achieve firstly a good representation and then a clustering (e.g. \cite{scan}).
In most cases, this unsupervised training is achieved by generating its own labels, and therefore the methods are called self-supervised.
A counterexample for an unsupervised method without self-supervision would be k-means \cite{k-means}.
Often, self-supervision is achieved on a pretext task on the same or a different dataset and then the pretrained network is fine-tuned on a downstream task \cite{survey-self}.
Many methods that follow this paradigm say their method is a form of representation learning \cite{simclr,self-context,self-jigsaw,self-jigsaw++,self-rotation}.
In this survey, we focus on image classification, and therefore most self-supervised or representation learning methods need to fine-tune on labeled data.
The combination of pretraining and fine-tuning can neither be called unsupervised nor self-supervised as external labeled information are used.
Semi-supervised learning describes methods that use labeled and unlabeled data. 
However, semi-supervised methods like \cite{fixmatch,remixmatch,mixmatch,uda,pseudolabel,mean-teacher,temporal-ensembling} use the labeled and unlabeled data from the beginning in comparison to representation learning methods like \cite{simclr,self-jigsaw,self-rotation,self-jigsaw++,self-context} which use them in different stages of their training.
Some methods combine ideas from self-supervised learning, semi-supervised learning and unsupervised learning \cite{S4L,foc} and are even more difficult to classify.

From the above explanation, we see that most methods are either unsupervised or semi-supervised in the context of image classification. 
The usage of labeled and unlabeled data in semi-supervised methods varies and a clear distinction in the common taxonomy is not obvious.
Nevertheless, we need to structure the methods in some way to keep an overview, allow comparisons and acknowledge the difference of research foci.
We decided against providing a fine-grained taxonomy as in previous literature \cite{survey-semi-unsuper} because we believe future research will come up with new combinations that were not thought of before.
We separate the methods only based on a rough distinction when the labeled or unlabeled data is used during the training.
For detailed comparisons, we distinct the methods based on their common ideas that are defined above and described in detail in \autoref{subsec:ideas}.
We call all semi-, self-, and unsupervised (learning) strategies together \emph{reduced supervised} (learning) strategies.

We defined \emph{stages} above (see \autoref{sec:pre}) as the different phases/time intervals during training when the different learning strategies supervised ($X = X_l$), unsupervised ($X = X_u$) or semi-supervised ($X_u \cap X_l \neq \emptyset$) are used.
For example, a method that uses a self-supervised pretraining on $X_u$ and then fine-tunes on the same images with labels has two stages.
A method that uses different algorithms, losses, or datasets during the training but only uses unsupervised data $X_u$ has one stage (e.g. \cite{scan}).
A method which uses $X_u$ and $X_l$ during the complete training has one stage (e.g. \cite{fixmatch}).
Based on the definition of stages during training, we classify reduced supervised methods into the training strategies: One-Stage-Semi-Supervised, One-Stage-Unsupervised, and Multi-Stage-Semi-Supervised.
An overview of the stage combinations and the corresponding training strategy is given in \autoref{fig:stages}.
As we concentrate on reduced supervised learning in this survey, we will not discuss any methods which are completely supervised.

Due to the above definition of stages a fifth combination of data usage between the stages exists.
This combination would use only labeled data in the first stage and unlabeled data in the second stage.
In the rest of the survey, we will exclude this training strategy for the following reasons. 
The case that a stage of complete supervision is followed by a stage of partial or no supervision is an unusual training strategy.
Due to this unusual usage, we only know of weight initialization followed by other reduced supervised training steps where this combination could occur.
We see the initialization of a network with pretrained weights from a supervised training on a different dataset (e.g. Imagenet \cite{imagenet}) as an architectural decision.
It is not part of the reduced supervised training process because it is used mainly as a more sophisticated weight initialization.
If we exclude weight initialization for this reason, we know of no method which belongs to this stage.

In the following paragraphs, we will describe all other training strategies in detail and they are illustrated in \autoref{fig:learning-strategies}.

\pic{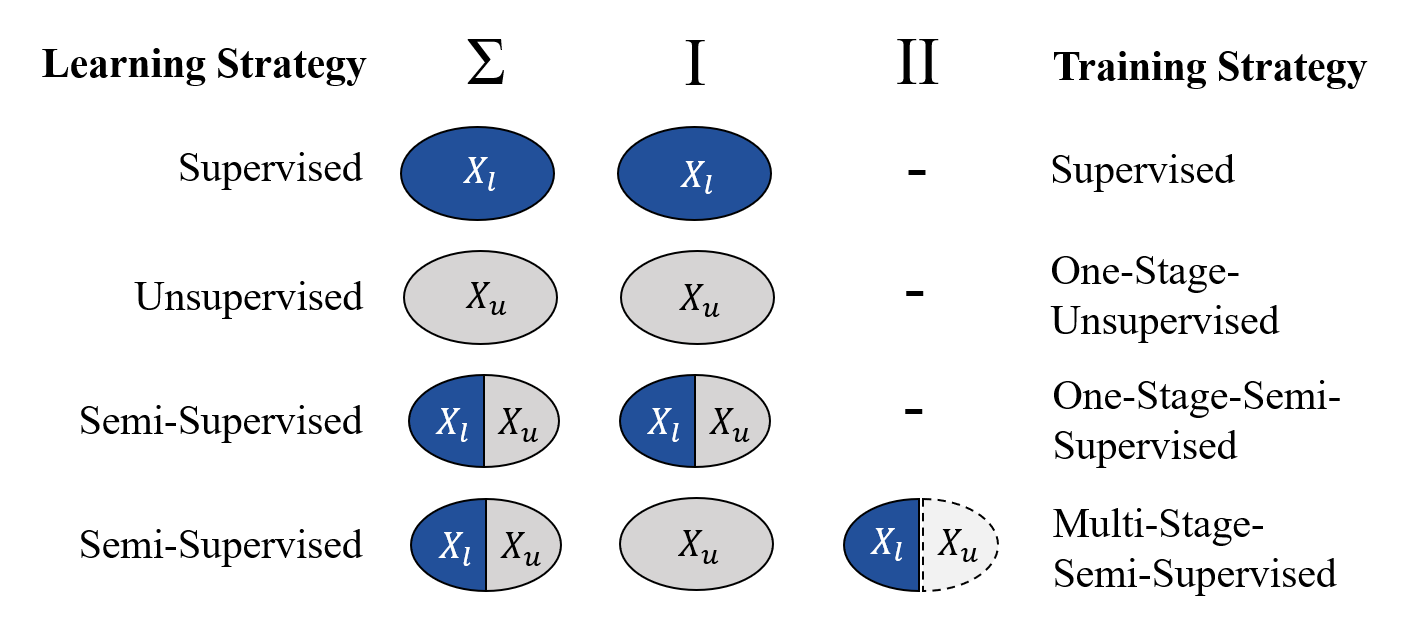}{Illustration of the different training strategies --
	Each row stands for a different combination of data usage during the first and second stage (defined in \autoref{sec:pre}).
	The first column states the common learning strategy name in the literature for this usage whereas the last column states the training strategy name used in this survey.
	The second column represents the used data overall. 
	The third and fourth column represent the used data in stage one or two. 
	The blue and grey (half-) circles represent the usage of the labeled data $X_l$ and the unlabeled data $X_u$ respectively in each stage or overall.
	A minus means that no further stage is used.
	The dashed half circle in the last row represents that this dashed part of the data can be used.
}

\subsubsection{Supervised Learning}

Supervised learning is the most common strategy in image classification with deep neural networks.
These methods only use labeled data $X_l$ and its corresponding labels $Z$.
The goal is to minimize a loss function between the output of the network $f(x)$ and the expected label $z_x \in Z$ for all $x \in X_l$.

\subsubsection{One-Stage-Semi-Supervised Training}

All methods which follow the one-stage-semi-supervised training strategy are trained in one stage with the usage of $X_l, X_u$, and $Z$.
The main difference to all supervised learning strategies is the usage of the additional unlabeled data $X_u$.
A common way to integrate the unlabeled data is to add one or more unsupervised losses to the supervised loss.

\subsubsection{One-Stage-Unsupervised Training}

All methods which follow the one-stage-unsupervised training strategy are trained in one stage with the usage of only the unlabeled samples $X_u$.
Therefore, many authors in this training strategy call their method unsupervised.
A variety of loss functions exist for unsupervised learning \cite{dac,iic,dec}.
In most cases, the problem is rephrased in such a way that all inputs for the loss can be generated, e.g. reconstruction loss in autoencoders \cite{dec}.
Due to this self-supervision, some call also these methods self-supervised.
We want to point out one major difference to many self-supervised methods following the multi-stage-semi-supervised training strategy below.
One-Stage-Unsupervised methods give image classifications without any further usage of labeled data.

\subsubsection{Multi-Stage-Semi-Supervised Training}

All methods which follow the multi-stage-semi-supervised training strategy are trained in two stages with the usage of $X_u$ in the first stage and $X_l$ and maybe $X_u$ in the second stage.
Many methods that are called self-supervised by their authors fall into this strategy.
Commonly a pretext task is used to learn representations on unlabeled data $X_u$.
In the second stage, these representations are fine-tuned to image classification on $X_l$. 
An important difference to a one-stage method is that these methods return useable classifications only after an additional training stage.

\subsection{Common ideas}
\label{subsec:ideas}

Different common ideas are used to train models in semi-, self-, and unsupervised learning. 
In this section, we present a selection of these ideas that are used across multiple methods in the literature.

It is important to notice that our usage of common ideas is fuzzy and incomplete by definition. 
A common idea should not be an identical implementation or approximation but the underlying motivation. 
This fuzziness is needed for two reasons. 
Firstly, a comparison would not be possible due to so many small differences in the exact implementations.
Secondly, they allow us to abstract some core elements of a method and therefore similarities can be detected.
Also, not all details, concepts, and motivations are captured by common ideas. 
We will limit ourselves to the common ideas described below since we believe they are enough to characterize all recent methods.
At the same time,  we know that these ideas need to be extended in the future as new common ideas will arise, old ones will disappear, and focus will shift to other ideas.
In contrast to detailed taxonomies, these new ideas can easily be integrated as new tags.

We sorted the ideas in alphabetical order and distinguish loss functions and general concepts.
Since ideas might reference each other, you may have to jump to the corresponding entry if you would like to know more.

\subsubsection*{\textbf{Loss Functions}}

\subsubsection*{Cross-entropy (CE)}
A common loss function for image classification is cross-entropy \cite{book-deep}.
It is commonly used to measure the difference between $f(x)$ and the corresponding label $z_x$ for a given $x \in X_l$.
The loss is defined in \autoref{eq:ce} and the goal is to minimize the difference.

\begin{equation}
\label{eq:ce}
\begin{split}
CE(z_x,f(x)) &= - \sum_{c=1}^{C} P(c| z_x)  log(P(c| f(x))) \\
&=  - \sum_{c=1}^{C} P(c| z_x)  log(P(c| z_x)) \\
&  -  \sum_{c=1}^{C} P(c| z_x)  log(\frac{P(c| f(x))}{P(c| z_x)})\\
&=  H(P(\cdot | z_x)) \\
& + KL(P(\cdot | z_x) ~||~ P( \cdot | f(x))
\end{split}
\end{equation}

$P$ is a probability distribution over all classes and is approximated with the (softmax-)output of the neural network $f(x)$ or the given label $z_x$.
$H$ is the entropy of a probability distribution and $KL$ is the Kullback-Leibler divergence.
It is important to note that cross-entropy is the sum of entropy over $z_x$ and a Kullback-Leibler divergence between $f(x)$ and $z_x$.
In general, the entropy $H(P(\cdot | z_x))$ is zero due to the one-hot encoded label $z_x$.

The loss function CE could also be used with a different probability distribution than $P$ based on the ground-truth label.
These distributions could be for example be based on Pseudo-Labels or other targets in a self-supervised pretext task.
We abbreviate the used common idea with CE* if not the ground-truth labels are used to highlight this specialty.

\picFour{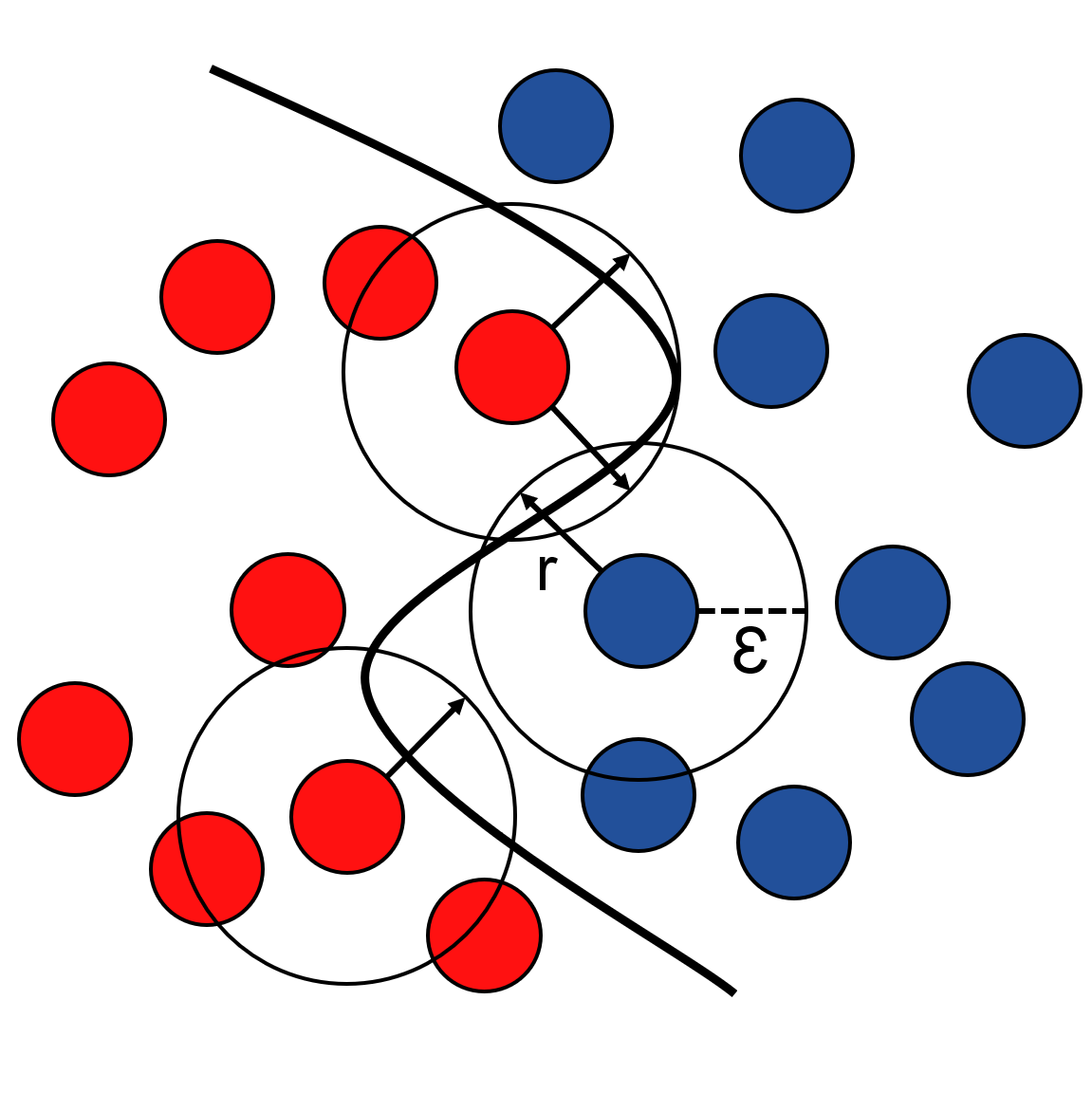}{VAT}{Mixup}{Overclustering}{Pseudo-Label}{Illustration of four selected common ideas --
	(a) The blue and red circles represent two different classes.
	The line is the decision boundary between these classes.
	The $\epsilon$ spheres around the circles define the area of possible transformations.
	The arrows represent the adversarial change vector $r$ which pushes the decision boundary away from any data point.
	(b) The images of a cat and a dog are combined with a parametrized blending.
	The labels are also combined with the same parameterization. 
	The shown images are taken from the dataset STL-10 \cite{stl-10}
	(c) Each circle represents a data point and the coloring of the circle the ground-truth label.
	In this example, the images in the middle have fuzzy ground-truth labels. 
	Classification can only draw one arbitrary decision boundary (dashed line) in the datapoints whereas overclustering can create multiple subregions. 
	This method could also be applied to outliers rather than fuzzy labels.
	(d) This loop represents one version of Pseudo-Labeling.
	A neural network predicts an output distribution. 
	This distribution is cast into a hard Pseudo-Label which is then used for further training the neural network.
}

\subsubsection*{Contrastive Loss (CL)}

A contrastive loss tries to distinguish positive and negative pairs.
The positive pair could be different views of the same image and the negative pairs could be all other pairwise combinations in a batch \cite{simclr}.
Hadsell \etal proposed to learn representations based on contrasting \cite{begin-contrastive}.
In recent years, the idea has been extended by self-supervised visual representation learning methods \cite{simclr,cmc,cpc,cpcv2,simclrv2}.
Examples of contrastive loss functions are NT-Xent \cite{simclr} and InfoNCE \cite{cpc} and both are based on Cross-Entropy.
The loss NT-Xent is computed across all positive pairs ($x_i,x_j$) in a fixed subset of $X$ with $N$ elements e.g. a batch during training.
The definition of the loss for a positive pair is given in \autoref{eq:ntxent}.
The similarity $sim$ between the outputs is measured with a normalized dot product, $\tau$ is a temperature parameter and the batch consists of $N$ image pairs.

\begin{equation}
\label{eq:ntxent}
\begin{split}
l_{x_i,x_j} = - log \frac{exp(sim(f(x_i),f(x_j)) / \tau)}{\sum_{k=1}^{2N} \mathds{1}_{k \neq i} exp(sim(f(x_i),f(x_k)) / \tau)}
\end{split}
\end{equation}

Chen and Li generalize the loss NT-Xent into a broader family of loss functions with an alignment and a distribution part \cite{interesting-constrastive}.
The alignment part encourages representations of positive pairs to be similar whereas the distribution part "encourages representations to match a prior distribution" \cite{interesting-constrastive}.
The loss InfoNCE is motivated like other contrastive losses by maximizing the agreement / mutual information between different views.
Van der Oord \etal showed that InfoNCE is a lower bound for the mutual information between the views \cite{cpc}.
More details and different bounds for other losses can be found in \cite{bounds-mi}.
However, Tschannen \etal show evidence that these lower bounds might not be the main reason for the successes of these methods \cite{self-supervised-mi}.
Due to this fact, we count losses like InfoNCE as a mixture of the common ideas contrastive loss and mutual information.

\subsubsection*{Entropy Minimization (EM)}
Grandvalet and Bengio noticed that the distributions of predictions in semi-supervised learning tend to be distributed over many or all classes instead of being sharp for one or few classes \cite{entropy-min}.
They proposed to sharpen the output predictions or in other words to force the network to make more confident predictions by minimizing entropy \cite{entropy-min}.
They minimized the entropy $H(P(\cdot | f(x)))$ for a probability distribution $(P(\cdot | f(x))$ based on a certain neural output $f(x)$ and an image $x \in X$.
This minimization leads to sharper / more confident predictions.
If this loss is used as the only loss the network/predictions would degenerate to a trivial minimization.

\subsubsection*{Kullback-Leibler divergence (KL)}
The Kullback-Leiber divergence is also commonly used in image classification since it can be interpreted as a part of cross-entropy.
In general, KL measures the difference between two given distributions \cite{kullback} and is therefore often used to define an auxiliary loss between the output $f(x)$ for an image $x \in X$ and a given secondary discrete probability distribution $Q$ over the classes $C$.
The definition is given in \autoref{eq:kull}. 
The second distribution could be another network output distribution, a prior known distribution, or a ground-truth distribution depending on the goal of the minimization.

\begin{equation}
\label{eq:kull}
\begin{split}
KL(Q ~||~ P( \cdot | f(x)) &= - \sum_{c=1}^{C} Q(c) log(\frac{P(c| f(x))}{Q(c)})
\end{split}
\end{equation}

\subsubsection*{Mean Squared Error (MSE)}
MSE measures the Euclidean distance  between two vectors e.g. two neural network outputs $f(x), f(y)$ for the images $x,y \in X$.
In contrast to the loss CE or KL, MSE is not a probability measure and therefore the vectors can be in an arbitrary Euclidean feature space (see \autoref{eq:mse}).
The minimization of the MSE will pull the two vectors or as in the example the network outputs together.
Similar to the minimization of entropy, this would lead to a degeneration of the network if this loss is used as the only loss on the network outputs.

\begin{equation}
\label{eq:mse}
MSE(f(x), f(y)) = || f(x) - f(y) || _2^2
\end{equation}

\subsubsection*{Mutual Information (MI)}
MI is defined for two probability distributions $P, Q$ as the Kullback Leiber (KL) divergence between the joint distribution and the marginal distributions \cite{ig}.
In many reduced supervised methods, the goal is to maximize the mutual information between the distributions. 
These distributions could be based on the input, the output, or an intermediate step of a neural network.
In most cases, the conditional distribution between $P$ and $Q$ and therefore the joint distribution is not known.
For example, we could use the outputs of a neural network $f(x), f(y)$ for two augmented views $x,y$ of the same image as the distributions $P, Q$.
In general, the distributions could be dependent as $x, y$ could be identical or very similar and the distributions could be independent if $x,y$ they are crops of distinct classes e.g. the background sky and the foreground object.
Therefore, the mutual information needs to be approximated.
The used approximation varies depending on the method and the definition of the distributions $P, Q$.
For further theoretical insights and several approximations see \cite{bounds-mi,mine}.

We show the definition of the mutual information between two network outputs $f(x),f(y)$ for images $x,y \in X$ as an example in \autoref{eq:mi}.
This equation also shows an alternative representation of mutual information: the separation in entropy $H(P( \cdot | f(x)))$ and conditional entropy $H(P( \cdot | f(x)) ~|~ P( \cdot | f(y)))$. 
Ji \etal argue that this representation illustrates the benefits of using MI over CE in unsupervised cases \cite{iic}.
A degeneration is avoided because MI balances the effects of maximizing the entropy with a uniform distribution for $P( \cdot | f(x))$ and minimizing the conditional entropy by equalizing $P( \cdot | f(x))$ and $P( \cdot | f(y))$.
Both cases lead to a degeneration of the neural network on their own.
\begin{equation}
\label{eq:mi}
\begin{split}
& I(P( \cdot | f(x),P( \cdot | f(y)) \\
& = KL(P( \cdot | f(x),f(y)) ~||~ P( \cdot | f(x) * P( \cdot | f(y)))) \\
& =  \sum_{c=1,c'=1}^{C}  P( c,c' | f(x),f(y)) \\
& ~~~~ ~~~~~ log(\frac{P( c,c' | f(x),f(y))}{P( c | f(x) * P( c' | f(y)))}) \\
& = H(P( \cdot | f(x)) + H(P( \cdot | f(x)) ~|~ P( \cdot | f(y)))
\end{split}
\end{equation}

\picEight{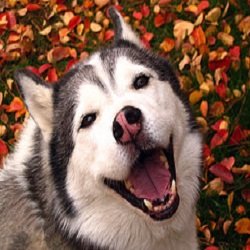}{Main image}{Different image}{Jigsaw}{Jigsaw++}{Exemplar}{Rotation}{Context}{Illustrations of 8 selected pretext tasks -- 
	(a) Example image for the pretext task
	(b) Negative/different example image in the dataset or batch
	(c) The Jigsaw pretext task consists of solving a simple Jigsaw puzzle generated from the main image.
	(d) Jigsaw++ augments the Jigsaw puzzle by adding in parts of a different image. 
	(e) In the exemplar pretext task, the distributions of a weakly augmented image (upper right corner) and several strongly augmented images should be aligned.
	(f) An image is rotated around a fixed set of rotations e.g. 0, 90, 180, and 270 degrees.
	The network should predict the rotation which has been applied.
	(g) A central patch and an adjacent patch from the same image are given.
	The task is to predict one of the 8 possible relative positions of the second patch to the first one.
	In the example, the correct answer is upper center.
	(h) The network receives a list of pairs and should predict the positive pairs.
	In this example, a positive pair consists of augmented views from the same image.
	Some illustrations are inspired by \cite{self-jigsaw++,self-context,self-rotation}.
}

\subsubsection*{Virtual Adversarial Training (VAT)}
VAT \cite{vat} tries to make predictions invariant to small transformations by minimizing the distance between an image and a transformed version of the image.
Miyato \etal showed how a transformation can be chosen and approximated in an adversarial way.
This adversarial transformation maximizes the distance between an image and a transformed version of it over all possible transformations.
The loss is defined in \autoref{eq:vat} with an image $x \in X$ and the output of a given neural network $f(x)$.
\begin{equation}
\label{eq:vat}
\begin{split}
VAT(f(x)) &=  D( P( \cdot | f(x) , P( \cdot | f(x + r_{adv}) ) \\
r_{adv} &= \argmax_{r;||r|| \leq \epsilon} D( P( \cdot | f(x) ,P( \cdot | f(x + r)) 
\end{split}
\end{equation}
$P$ is the probability distribution over the outputs of the neural network and $D$ is a non-negative function that measures the distance.
As illustrated in \autoref{fig:method-0} r is a vector and $\epsilon$ the maximum length of this vector.
Two examples of used distance measures are cross-entropy \cite{vat} and Kullback-Leiber divergence \cite{S4L}.

\subsubsection*{\textbf{Concepts}}

\subsubsection*{Mixup (MU)}
Mixup creates convex combinations of images by blending them into each other.
An illustration of the concept is given in \autoref{fig:method-1}.
The prediction of the convex combination of the corresponding labels turned out to be beneficial because the network needs to create consistent predictions for intermediate interpolations of the image.
This approach has been beneficial for supervised learning in general \cite{mixup} and is therefore also used in several semi-supervised learning algorithms \cite{mixmatch,remixmatch,fixmatch}.

\subsubsection*{Overclustering (OC)}
Normally, if we have $k$ classes in the supervised case we also use $k$ clusters in the unsupervised case.
Research showed that it can be beneficial to use more clusters than actual classes $k$ exist \cite{deep-cluster,iic,foc}.
We call this idea \emph{overclustering}.
Overclustering can be beneficial in semi-supervised or unsupervised cases due to the effect that neural networks can decide 'on their own' how to split the data.
This separation can be helpful in noisy/fuzzy data or with intermediate classes that were sorted into adjacent classes randomly \cite{foc}.
An illustration of this idea is presented in \autoref{fig:method-2}

\subsubsection*{Pretext Task (PT)}
A pretext task is a broad-ranged description of self-supervised training a neural network on a different task than the target task.
This task can be for example predicting the rotation of an image \cite{self-rotation}, solving a jigsaw puzzle \cite{self-jigsaw}, using a contrastive loss \cite{simclr,cpc} or maximizing mutual information \cite{iic,foc}.
An overview of most pretext task in this survey is given in \autoref{fig:pretext} and a complete overview is given in \autoref{tab:methods}.
In most cases the self-supervised, pretext task is used to learn representations which can then be fine-tuned for image classification \cite{simclr,cpc,self-exemplar,self-context,self-jigsaw,self-jigsaw++,self-rotation}.
In a semi-supervised context, some methods use this pretext task to define an additional loss during training \cite{remixmatch}.

\subsubsection*{Pseudo-Labels (PL)}
A simple approach for estimating labels of unknown data is using Pseudo-Labels \cite{pseudolabel}.
Lee proposed to classify unseen data with a neural network and use the predictions as labels.
This process is illustrated in \autoref{fig:method-3}.
What sounds at first like a self-fulfilling assumption works reasonably well in real-world image classification tasks. 
It is important to notice that the network needs additional information to prevent total random predictions.
This additional information could be some known labels or a weight initialization of other supervised data or unsupervised on a pretext task.
Several modern methods are based on the same core idea of creating labels by predicting them on their own \cite{mean-teacher,mixmatch}.

\section{Methods}
\label{sec:methods}

This section shorty summarizes all methods in the survey in roughly chronological order and separated by their training strategy.
Each summary states the used common ideas, explains their usage, and highlights special cases.
The abbreviations for the common ideas are defined in \autoref{subsec:ideas}.
We include a large number of recent methods but we do not claim this list to be complete.

\subsection{One-Stage-Semi-Supervised}

\picFour{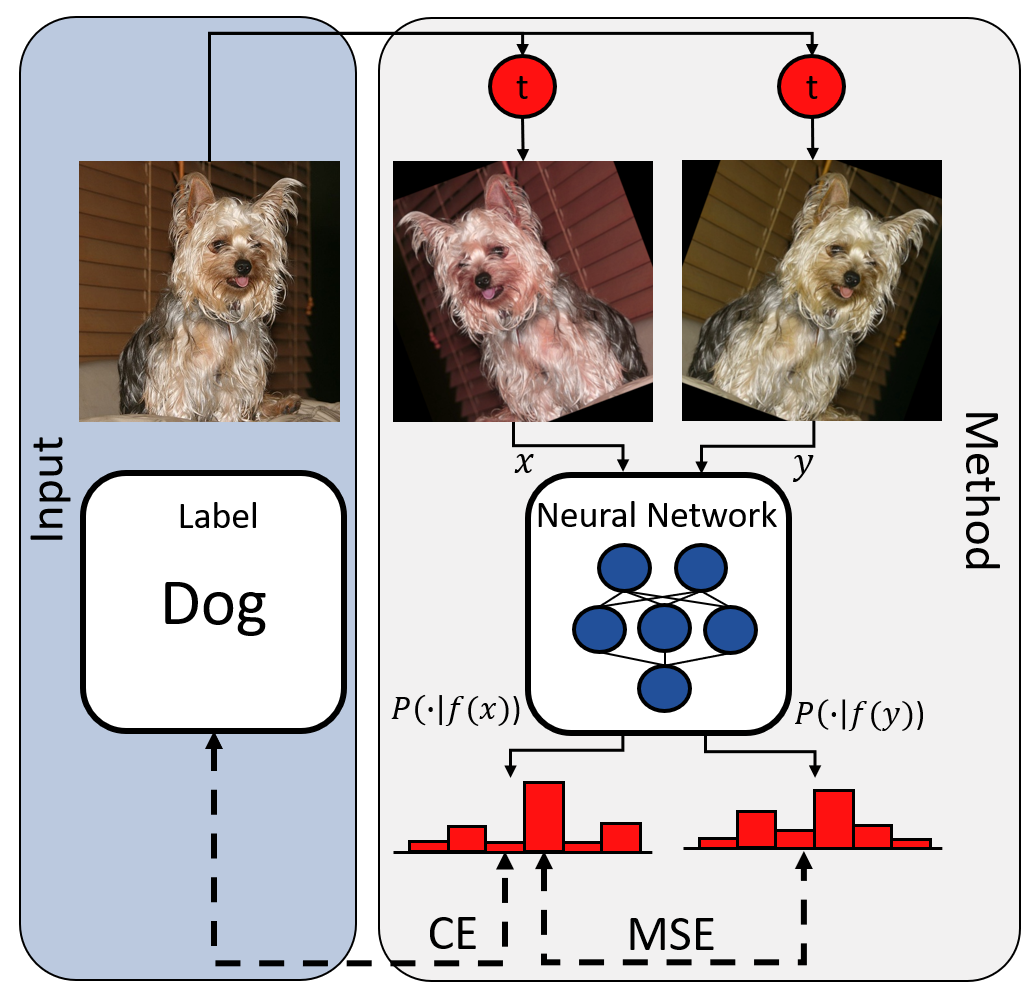}{$\pi$-model}{Temporal Ensembling}{Mean Teacher}{UDA}{Illustration of four selected one-stage-semi-supervised methods -- 
	The used method is given below each image.
	The input including label information is given in the blue box on the left side.
	On the right side, an illustration of the method is provided.
	In general, the process is organized from top to bottom. 
	At first, the input images are preprocessed by none or two different random transformations $t$.
	Special augmentation techniques like Autoaugment \cite{autoaugment} are represented by a red box.
	The following neural network uses these preprocessed images ($x,y$) as input.
	The calculation of the loss (dotted line) is different for each method but shares common parts.
	All methods use the cross-entropy (CE) between label and predicted distribution $P(\cdot |f(x))$ on labeled examples.
	Details about the methods can be found in the corresponding entry in \autoref{sec:methods} whereas abbreviations for common methods are defined in \autoref{subsec:ideas}.
	EMA stands for the exponential moving average.
} 

\subsubsection*{Pseudo-Labels} 
Pseudo-Labels \cite{pseudolabel} describes a common idea in deep learning and a learning method on its own.
For the description of the common idea see above in \autoref{subsec:ideas}.
In contrast to many other semi-supervised methods, Pseudo-Labels does not use a combination of an unsupervised and a supervised loss.
The Pseudo-Labels approach uses the predictions of a neural network as labels for unknown data as described in the common idea.
Therefore, the labeled and unlabeled data are used in parallel to minimize the CE loss.
\emph{Common ideas: CE, CE*, PL}

\subsubsection*{$\pi$-model and Temporal Ensembling}
Laine \& Aila present two similar learning methods with the names $\pi$-model and Temporal Ensembling \cite{temporal-ensembling}.
Both methods use a combination of the supervised CE loss and the unsupervised consistency loss MSE.
The first input for the consistency loss in both cases is the output of their network from a randomly augmented input image.
The second input is different for each method.
In the $\pi$-model an augmentation of the same image is used.
In Temporal Ensembling an exponential moving average of previous predictions is evaluated.
Laine \& Aila show that Temporal Ensembling is up to two times faster and more stable in comparison to the $\pi$-model \cite{temporal-ensembling}.
Illustrations of these methods are given in \autoref{fig:models}.
\emph{Common ideas: CE, MSE}

\subsubsection*{Mean Teacher}
With Mean Teacher Tarvainen \& Valpola present a student-teacher-approach for semi-supervised learning \cite{mean-teacher}.
They develop their approach based on the $\pi$-model and Temporal Ensembling \cite{temporal-ensembling}.
Therefore, they also use MSE as a consistency loss between two predictions but create these predictions differently.
They argue that Temporal Ensembling incorporates new information too slowly into predictions.
The reason for this is that the exponential moving average (EMA) is only updated once per epoch.
Therefore, they propose to use a teacher based on the average weights of a student in each update step.
Tarvainen \& Valpola show for their model that the KL-divergence is an inferior consistency loss than MSE.
An illustration of this method is given in \autoref{fig:models}.
\emph{Common ideas: CE, MSE}

\subsubsection*{Virtual Adversarial Training (VAT)}
VAT \cite{vat} is not just the name for a common idea but it is also a one-stage-semi-supervised method.
Miyato \etal used a combination of VAT on unlabeled data and CE on labeled data \cite{vat}.
They showed that the adversarial transformation leads to a lower error on image classification than random transformations.
Furthermore, they showed that adding EntMin \cite{entropy-min} to the loss increased accuracy even more.
\emph{Common ideas: CE, (EM), VAT}

\subsubsection*{Interpolation Consistency Training (ICT)}\mbox{} \\
ICT \cite{ict} uses linear interpolations of unlabeled data points to regularize the consistency between images.
Verma \etal use a combination of the supervised loss CE and the unsupervised loss MSE.
The unsupervised loss is measured between the prediction of the interpolation of two images and the interpolation of their Pseudo-Labels.
The interpolation is generated with the mixup \cite{mixup} algorithm from two unlabeled data points.
For these unlabeled data points, the Pseudo-Labels are predicted by a Mean Teacher \cite{mean-teacher} network.
\emph{Common ideas: CE, MSE, MU, PL}

\subsubsection*{Fast-Stochastic Weight Averaging (fast-SWA)}
In contrast to other semi-supervised methods, Athiwaratkun \etal do not change the loss but the optimization algorithm \cite{fast-SWA}.
They analyzed the learning process based on ideas and concepts of SWA \cite{SWA}, $\pi$-model \cite{temporal-ensembling} and Mean Teacher \cite{mean-teacher}.
Athiwaratkun \etal show that averaging and cycling learning rates are beneficial in semi-supervised learning by stabilizing the training.
They call their improved version of SWA fast-SWA due to faster convergence and lower performance variance \cite{fast-SWA}.
The architecture and loss is either copied from $\pi$-model \cite{temporal-ensembling} or Mean Teacher \cite{mean-teacher}.
\emph{Common ideas: CE, MSE}

\picFour{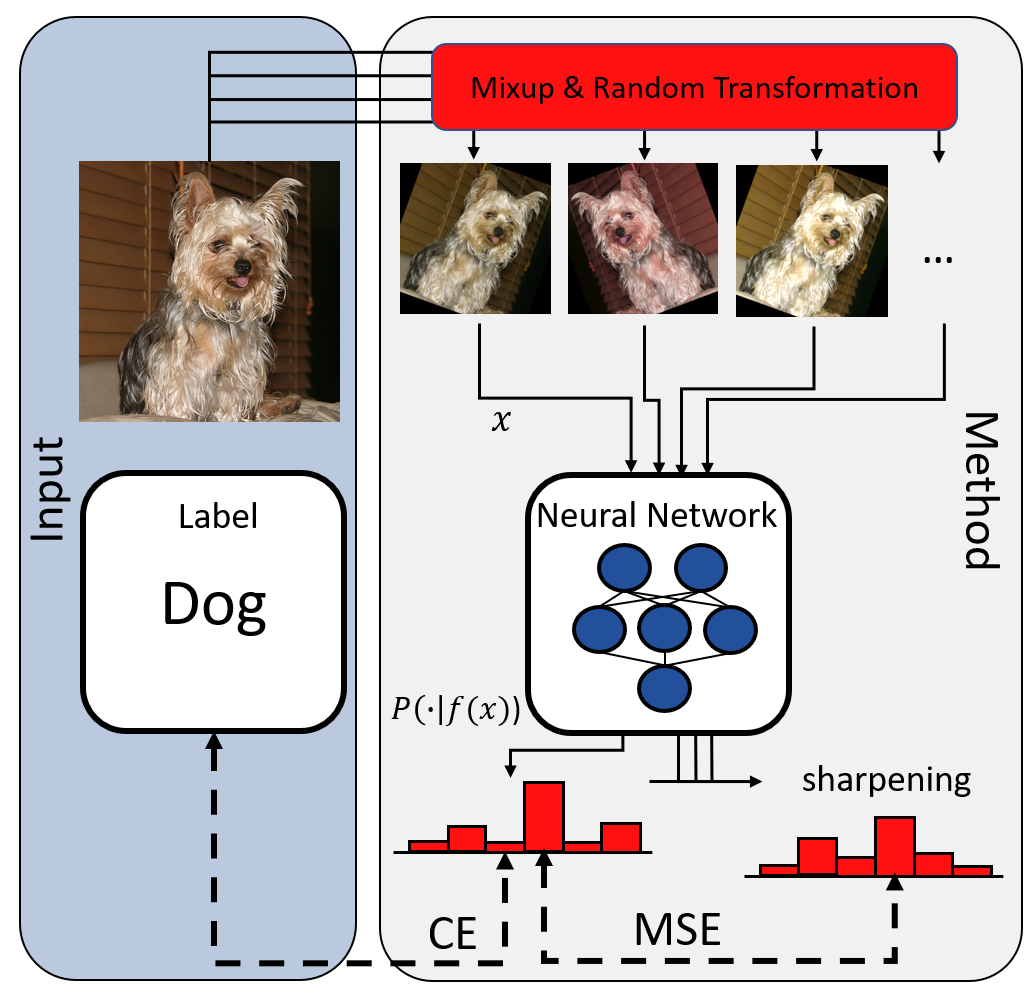}{MixMatch}{ReMixMatch}{FixMatch}{FOC}{Illustration of four selected methods -- 
	The used method is given below each image.
	The input including label information is given in the blue box on the left side.
	On the right side, an illustration of the method is provided.
	For FOC the second stage is represented.
	In general, the process is organized from top to bottom. 
	At first, the input images are preprocessed by none or two different random transformations $t$.
	Special augmentation techniques like CTAugment \cite{remixmatch} are represented by a red box.
	The following neural network uses these preprocessed images (e.g. $x,y$) as input.
	The calculation of the loss (dotted line) is different for each method but shares common parts.
	All methods use the cross-entropy (CE) between label and predicted distribution $P(\cdot |f(x))$ on labeled examples.
	Details about the methods can be found in the corresponding entry in \autoref{sec:methods} whereas abbreviations for common methods are defined in \autoref{subsec:ideas}.
} 

\subsubsection*{MixMatch} 
MixMatch \cite{mixmatch} uses a combination of a supervised and an unsupervised loss.
Berthelot \etal use CE as the supervised loss and MSE between predictions and generated Pseudo-Labels as their unsupervised loss.
These Pseudo-Labels are created from previous predictions of augmented images. 
They propose a novel sharping method over multiple predictions to improve the quality of the Pseudo-Labels.
This sharpening also enforces implicitly a minimization of the entropy on the unlabeled data.
Furthermore, they extend the algorithm mixup \cite{mixup} to semi-supervised learning by incorporating the generated labels.
\emph{Common ideas: CE, (EM), MSE, MU, PL}

\subsubsection*{Ensemple AutoEndocing Transformation (EnAET)}
EnAET \cite{enaet} combines the self-supervised pretext task AutoEncoding Transformations \cite{aet} with MixMatch \cite{mixmatch}.
Wang \etal apply spatial transformations, such as translations and rotations, and non-spatial transformations, such as color distortions, on input images in the pretext task. 
The transformations are then estimated with the original and augmented image given.
This is a difference to other pretext tasks where the estimation is often based on the augmented image only \cite{self-rotation}.
The loss is used together with the loss of MixMatch and is extended with the Kullback Leiber divergence between the predictions of the original and the augmented image. 
\emph{Common ideas: CE, (EM), KL, MSE, MU, PL, PT}

\subsubsection*{Unsupervised Data Augmentation (UDA)}
Xie \etal present with UDA a semi-supervised learning algorithm that concentrates on the usage of state-of-the-art augmentation \cite{uda}.
They use a supervised and an unsupervised loss.
The supervised loss is CE whereas the unsupervised loss is the Kullback Leiber divergence between output predictions.
These output predictions are based on an image and an augmented version of this image.
For image classification, they propose to use the augmentation scheme generated by AutoAugment \cite{autoaugment} in combination with Cutout \cite{cutout}.
AutoAugment uses reinforcement learning to create useful augmentations automatically.
Cutout is an augmentation scheme where randomly selected regions of the image are masked out.
Xie \etal show that this combined augmentation method achieves higher performance in comparison to previous methods on their own like Cutout, Cropping, or Flipping.
In addition to the different augmentation, they propose to use a variety of other regularization methods.
They proposed Training Signal Annealing which restricts the influence of labeled examples during the training process to prevent overfitting.
They use EntMin \cite{entropy-min} and a kind of Pseudo-Labeling \cite{pseudolabel}.
We use the term kind of Pseudo-Labeling because they do not use the predictions as labels but they use them to filter unsupervised data for outliers.
An illustration of this method is given in \autoref{fig:models}.
\emph{Common ideas: CE,  EM, KL, (PL)}

\subsubsection*{Self-paced Multi-view Co-training (SpamCo)}
Ma \etal propose a general framework for co-training across multiple views \cite{cotraining}.
In the context of image classification, different neural networks can be used as different views.
The main idea of the co-training between different views is similar to using Pseudo-Labels.
The main differences in SpamCo are that the Pseudo-Labels are not used for all samples and they influence each other across views.
Each unlabeled image has a weight value for each view. 
Based on an age parameter, more unlabeled images are considered in each iteration.
At first only confident Pseudo-Labels are used and over time also less confident ones are allowed.
The proposed hard or soft co-regularizers also influence the weighting of the unlabeled images.
The regularizers encourage to select unlabeled images for training across views. 
Without this regularization the training would degenerate to an independent training of the different views/models.
CE is used as loss on the labels and Pseudo-Labels with additional $L_2$ regularization.
Ma \etal show further applications including text classification and object detection.
\emph{Common ideas: CE, CE*, MSE, PL}

\subsubsection*{ReMixMatch}
ReMixMatch \cite{remixmatch} is an extension of MixMatch with distribution alignment and augmentation anchoring.
Berthelot \etal motivate the distribution alignment with an analysis of mutual information. 
They use entropy minimization via "sharpening" but they do not use any prediction equalization like in mutual information.
They argue that an equal distribution is also not desirable since the distribution of the unlabeled data could be skewed.
Therefore, they align the predictions of the unlabeled data with a marginal class distribution over the seen examples. 
Berthelot \etal exchange the augmentation scheme of MixMatch with augmentation anchoring.
Instead of averaging the prediction over different slight augmentations of an image they only use stronger augmentations as regularization. 
All augmented predictions of an image are encouraged to result in the same distribution with CE instead of MSE.
Furthermore, a self-supervised loss based on the rotation pretext task \cite{self-rotation} was added.
\emph{Common ideas: CE, CE* (EM), (MI), MU, PL, PT}

\subsubsection*{FixMatch}
FixMatch \cite{fixmatch} is building on the ideas of ReMixMatch but is dropping several ideas to make the framework more simple while achieving a better performance.
FixMatch is using the cross-entropy loss on the supervised and the unsupervised data. 
For each image in the unlabeled data, one weakly- and one strongly-augmented version is created.
The Pseudo-Label of the weakly-augmented version is used if a confidence threshold is surpassed by the network.
If a Pseudo-Label is calculated the network output of the strongly-augmented version is compared with this hard label via cross-entropy which implicitly encourages low-entropy predictions on the unlabeled data \cite{fixmatch}.
Sohn \etal do not use ideas like Mixup, VAT, or distribution alignment but they state that they can be used and provide ablations for some of these extensions.
\emph{Common ideas: CE, CE*, (EM),  PL}

\picFour{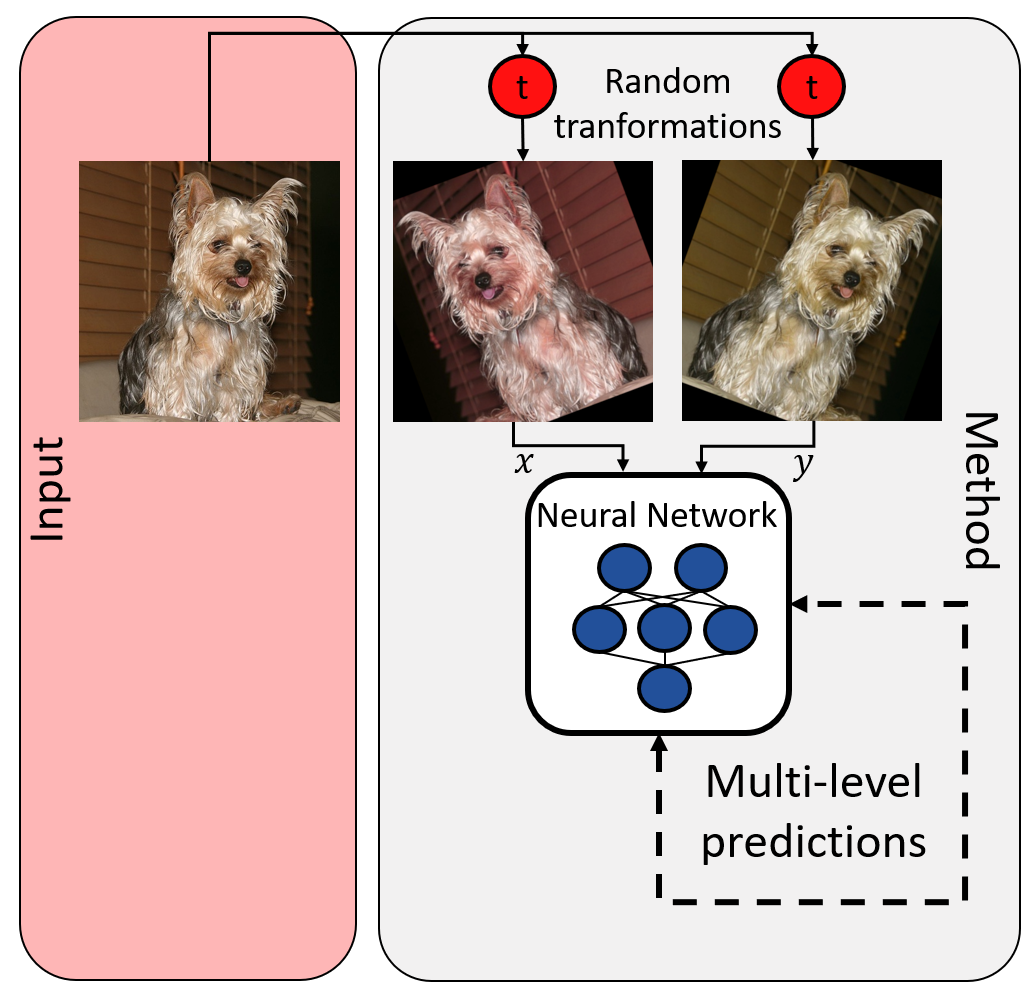}{AMDIM}{CPC}{DeepCluster}{IIC}{
	Illustration of four selected multi-stage-semi-supervised methods -- 
	The used method is given below each image.
	The input is given in the red box on the left side.
	On the right side, an illustration of the method is provided.
	The fine-tuning part is excluded and only the first stage/pretext task is represented.
	In general, the process is organized from top to bottom. 
	At first, the input images are either preprocessed by one or two random transformations $t$ or are split up.
	The following neural network uses these preprocessed images ($x,y$) as input.
	The calculation of the loss (dotted line) is different for each method.
	AMDIM and CPC use internal elements of the network to calculate the loss.
	DeepCluster and IIC use the predicted output distributions ($P( \cdot |f(x)),P( \cdot |f(y))$) to calculate a loss.
	Details about the methods can be found in the corresponding entry in \autoref{sec:methods} whereas abbreviations for common methods are defined in \autoref{subsec:ideas}.
}
\subsection{Multi-Stage-Semi-Supervised}

\subsubsection*{Exemplar}
Dosovitskiy \etal proposed a self-supervised pretext task with additional fine-tuning \cite{self-exemplar}.
They randomly sample patches from different images and augment these patches heavily.
Augmentations can be for example rotations, translations, color changes, or contrast adjustments.
The classification task is to map all augmented versions of a patch to the correct original patch using cross-entropy loss.
\emph{Common ideas: CE, CE*, PT}

\subsubsection*{Context}
Doersch \etal propose to use context prediction as a pretext task for visual representation learning \cite{self-context}.
A central patch and an adjacent patch from an image are used as input.
The task is to predict one of the 8 possible relative positions of the second patch to the first one using cross-entropy loss.
An illustration of the pretext task is given in \autoref{fig:pretext}.
Doersch \etal argue that this task becomes easier if you recognize the content of these patches.
The authors fine-tune their representations for other tasks and show their superiority in comparison to the random initialization.
Aside from fine-tuning, Doersch \etal show how their method could be used for Visual Data Mining. 
\emph{Common ideas: CE, CE*, PT}

\subsubsection*{Jigsaw}
Noroozi and Favaro propose to solve Jigsaw puzzles as a pretext task \cite{self-jigsaw}.
The idea is that a network has to understand the concept of a presented object to solve the puzzle using the classification loss cross-entropy.
They prevent simple solutions that only look at edges or corners by including small random margins between the puzzle patches.
They fine-tune on supervised data for image classification tasks.
Noroozi \etal extended the Jigsaw task by adding image parts of a different image \cite{self-jigsaw++}.
They call the extension Jigsaw++.
Examples for a Jigsaw or Jigsaw++ puzzle are given in \autoref{fig:pretext}.
\emph{Common ideas: CE, CE*, PT}

\subsubsection*{DeepCluster}
DeepCluster \cite{deep-cluster} is a self-supervised method that generates labels by k-means clustering.
Caron \etal iterate between clustering of predicted labels to generate Pseudo-Labels and training with cross-entropy on these labels.
They show that it is beneficial to use overclustering in the pretext task.
After the pretext task, they fine-tune the network on all labels.
An illustration of this method is given in \autoref{fig:models2}.
\emph{Common ideas: CE, OC, PL, PT}

\subsubsection*{Rotation}
Gidaris \etal use a pretext task based on image rotation prediction \cite{self-rotation}.
They propose to randomly rotate the input image by 0, 90, 180, or 270 degrees and let the network predict the chosen rotation degree.
They train the network with cross-entropy on this classification task.
In their work, they also evaluate different numbers of rotations but four rotations score the best result.
For image classification, they fine-tune on labeled data.
\emph{Common ideas: CE, CE*, PT}

\subsubsection*{Contrastive Predictive Coding (CPC)}
CPC \cite{cpc,cpcv2} is a self-supervised method that predicts representations of local image regions based on previous image regions.
The authors determine the quality of these predictions with a contrastive loss which identifies the correct prediction out of randomly sampled negative ones.
They call their loss InfoNCE which is cross-entropy for the prediction of positive examples \cite{cpc}.
Van den Oord \etal showed that minimizing InfoNCE maximizes the lower bound for MI between the previous image regions and the predicted image region \cite{cpc}.
An illustration of this method is given in \autoref{fig:models2}.
The representations of the pretext task are then fine-tuned.
\emph{Common ideas: CE, (CE*), CL, (MI), PT}

\subsubsection*{Constrastive Multiview Coding (CMC)}
CMC \cite{cmc} generalizes CPC \cite{cpc} to an arbitrary collection of views.
Tian \etal try to learn an embedding that is different for contrastive samples and equal for similar images. 
Like Oord \etal they train their network by identifying the correct prediction out of multiple negative ones \cite{cpc}.
However, Tian \etal take different views of the same image such as color channels, depth, and segmentation as similar images.
For common image classification datasets like STL-10, they use patch-based similarity.
After this pretext task, the representations are fine-tuned to the desired dataset. 
\emph{Common ideas: CE, (CE*), CL, (MI), PT}

\subsubsection*{Deep InfoMax (DIM)}
DIM \cite{dim} maximizes the MI between local input regions and output representations.
Hjelm \etal show that maximizing over local input regions rather than the complete image is beneficial for image classification.
Also, they use a discriminator to match the output representations to a given prior distribution.
In the end, they fine-tune the network with an additional small fully-connected neural network. 
\emph{Common ideas: CE, MI, PT}

\subsubsection*{Augmented Multiscale Deep InfoMax (AMDIM)}
AMDIM \cite{amdim} maximizes the MI between inputs and outputs of a network. 
It is an extension of the method DIM  \cite{dim}.
DIM usually maximizes MI between local regions of an image and a representation of the image.
AMDIM extends the idea of DIM in several ways.
Firstly, the authors sample the local regions and representations from different augmentations of the same source image.
Secondly, they maximize MI between multiple scales of the local region and the representation.
They use a more powerful encoder and define mixture-based representations to achieve higher accuracies.
Bachman \etal fine-tune the representations on labeled data to measure their quality.
An illustration of this method is given in \autoref{fig:models2}.
\emph{Common ideas: CE, MI, PT}

\subsubsection*{Deep Metric Transfer (DMT)}
DMT \cite{dmt} learns a metric as a pretext task and then propagates labels onto unlabeled data with this metric. 
Liu \etal use  self-supervised image colorization \cite{metric-color} or unsupervised instance discrimination \cite{metric-instance} to calculate a metric.
In the second stage, they propagate labels to unlabeled data with spectral clustering and then fine-tune the network with the new Pseudo-Labels.
Additionally, they show that their approach is complementary to previous methods.
If they use the most confident Pseudo-Labels for methods such as Mean Teacher \cite{mean-teacher} or VAT \cite{vat}, they can improve the accuracy with very few labels by about 30\%.
\emph{Common ideas: CE, CE*, PL, PT}

\subsubsection*{Invariant Information Clustering (IIC)}
IIC \cite{iic} maximizes the MI between augmented views of an image.
The idea is that images should belong to the same class regardless of the augmentation.
The augmentation has to be a transformation to which the neural network should be invariant.
The authors do not maximize directly over the output distributions but over the class distribution which is approximated for every batch.
Ji \etal use auxiliary overclustering on a different output head to increase their performance in the unsupervised case.
This idea allows the network to learn subclasses and handle noisy data.
Ji \etal use Sobel filtered images as input instead of the original RGB images.
Additionally, they show how to extend IIC to image segmentation.
Up to this point, the method is completely unsupervised. 
To be comparable to other semi-supervised methods they fine-tune their models on a subset of available labels.
An illustration of this method is given in \autoref{fig:models2}.
The first unsupervised stage can be seen as a self-supervised pretext task.
In contrast to other pretext tasks, this task already predicts representations which can be seen as classifications.
\emph{Common ideas: CE, MI, OC, PT}

\subsubsection*{Self-Supervised Semi-Supervised Learning (S$^4$L)} 
S$^4$L \cite{S4L} is, as the name suggests, a combination of self-supervised and semi-supervised methods.
Zhai \etal split the loss into a supervised and an unsupervised part.
The supervised loss is CE whereas the unsupervised loss is based on the self-supervised techniques using rotation and exemplar prediction \cite{self-rotation,self-exemplar}.
The authors show that their method performs better than other self-supervised and semi-supervised techniques \cite{self-exemplar,self-rotation, vat,entropy-min,pseudolabel}.
In their \textit{Mix Of All Models} (MOAM) they combine self-supervised rotation prediction, VAT, entropy minimization, Pseudo-Labels, and fine-tuning into a single model with multiple training steps.
Since we discuss the results of their MOAM we identify S$^4$L as a multi-stage-semi-supervised method.
\emph{Common ideas: CE, CE*, EM, PL, PT, VAT}

\picFour{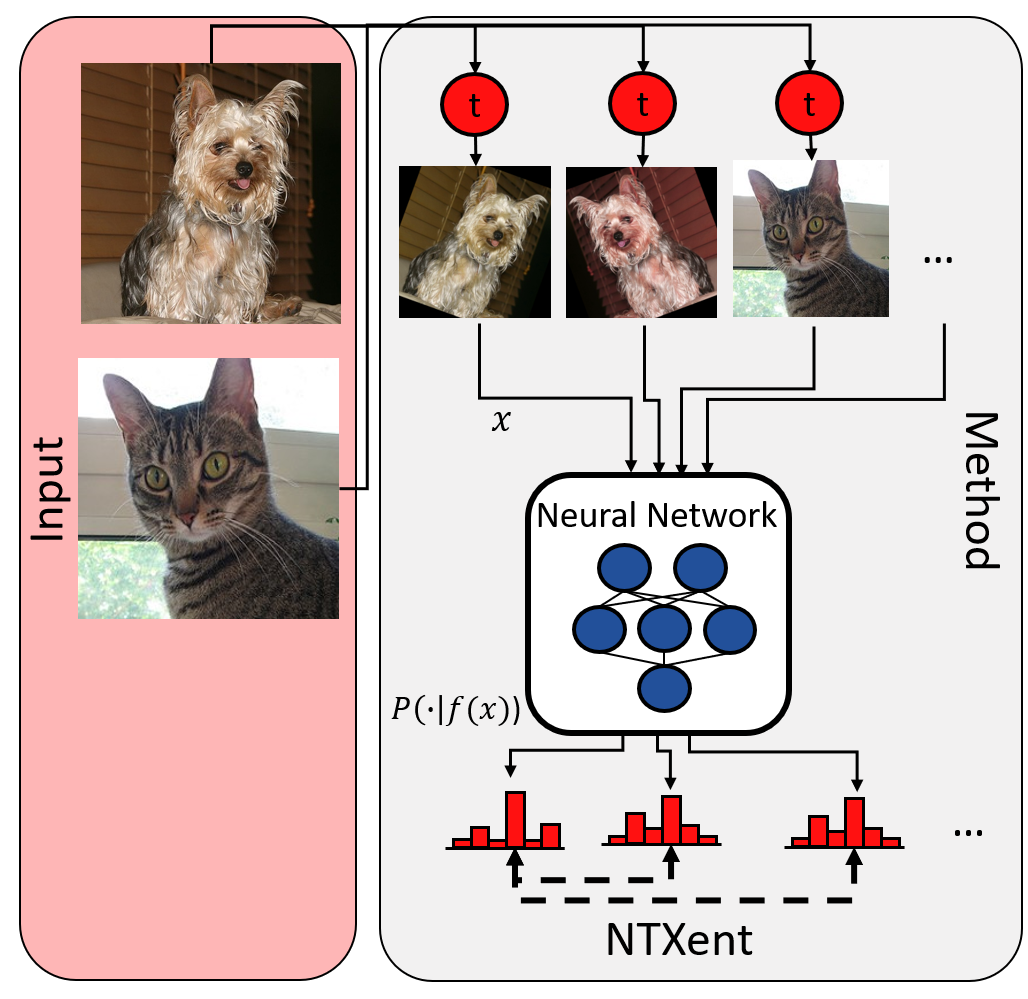}{SimCLR}{SimCLRv2}{MoCo}{BYOL}{
	Illustration of four selected multi-stage-semi-supervised methods -- 
	The used method is given below each image.
	The input is given in the red (not using labels) or blue (using labels) box on the left side.
	On the right side, an illustration of the method is provided.
	The fine-tuning part is excluded and only the first stage/pretext task is represented.
	For SimCLRv2 the second stage or distillation step is illustrated. 
	In general, the process is organized from top to bottom. 
	At first, the input images are either preprocessed by one or two random transformations $t$ or are split up.
	The following neural network uses these preprocessed images ($x,y$) as input.
	Details about the methods can be found in the corresponding entry in \autoref{sec:methods} whereas abbreviations for common methods are defined in \autoref{subsec:ideas}.
	EMA stands for the exponential moving average.
}

\subsubsection*{Simple Framework for Contrastive Learning of Visual Representation (SimCLR)}
SimCLR \cite{simclr} maximizes the agreement between two different augmentations of the same image.
The method is similiar to CPC \cite{cpc} and IIC \cite{iic}.
In comparison to CPC Chen \etal do not use the different inner representations.
Contrary to IIC they use normalized temperature-scaled cross-entropy (NT-Xent) as their loss.
Based on the cosine similarity of the predictions, NT-Xent measures whether positive pairs are similar and negative pairs are dissimilar.
Augmented versions of the same image are treated as positive pairs and pairs with any other image as negative pair.
The system is trained with large batch sizes of up to 8192 instead of a memory bank to create enough negative examples.
\emph{Common ideas: CE, (CE*), CL, PT}

\subsubsection*{Fuzzy Overclustering (FOC)}
Fuzzy Overclustering \cite{foc} is an extension of IIC \cite{iic}.
FOC focuses on using overclustering to subdivide fuzzy labels in real-world datasets. 
Therefore, it unifies the used data and losses proposed by IIC between the different stages and extends it with new ideas such as the novel loss Inverse Cross-Entropy (CE$^{-1}$).
This loss is inspired by Cross-Entropy but can be used on the overclustering results of the network where no ground truth labels are known.
FOC is not achieving state-of-the-art results on a common image classification dataset.
However, on a real-world plankton dataset with fuzzy labels, it surpasses FixMatch and shows that 5-10\% more consistent predictions can be achieved.
Like IIC, FOC can be viewed as a multi-stage-semi-supervised and an one-stage-unsupervised method.
In general, FOC is trained in one unsupervised and one semi-supervised stage and can be seen as a multi-stage-semi-supervised method.
Like IIC, it produces classifications already in the unsupervised stage and can therefore also be seen as an one-stage-unsupervised method.
\emph{Common ideas: CE, (CE*) MI, OC, PT}

\subsubsection*{Momentum Contrast (MoCo)}
He \etal  propose to use a momentum encoder for contrastive learning \cite{moco}.
In other methods \cite{simclr,simclrv2,cpc,cpcv2}, the negative examples for the contrastive loss are sampled from the same mini-batch as the positive pair. 
A large batch size is needed to ensure a great variety of negative examples.
He \etal sample their negative examples from a queue encoded by another network whose weights are updated with an exponential moving average of the main network.
They solve the pretext task proposed by \cite{metric-instance} with negative examples samples from their queue and fine-tune in a second stage on labeled data.
Chen \etal provide further ablations and baseline for the MoCo Framework e.g. by using a MLP head for fine-tuning \cite{mocov2}.
\emph{Common ideas: CE, CL, PT}

\subsubsection*{Bootstrap you own latent (BYOL)}
Grill \etal use an online and a target network.
In the proposed pretext task, the online network predicts the image representation of the target network for an image \cite{byol}.
The difference between the predictions is measured with MSE.
Normally, this approach would lead to a degeneration of the network as a constant prediction over all images would also achieve the goal.
In contrastive learning, this degeneration is avoided by selecting a positive pair of examples from multiple negative ones \cite{simclr,simclrv2,cpc,cpcv2,moco,mocov2}.
By using a slow-moving average of the weights between the online and target network, Grill \etal show empirically that the degeneration to a constant prediction can be avoided.
This approach has the positive effect that BYOL performance is depending less on hyperparameters like augmentation and batch size \cite{byol}.
In a follow-up work, Richemond \etal show that BYOL even works when no batch normalization which might have introduced kind of a contrastive learning effect in the batches is used  \cite{byolv2}.
\emph{Common ideas: MSE, PT}

\subsubsection*{Simple Framework for Contrastive Learning of Visual Representation (SimCLRv2)}
Chen \etal extend the framework SimCLR by using larger and deeper networks and by incorporating the memory mechanism from MoCo \cite{simclrv2}.
Moreover, they propose to use this framework in three steps. 
The first is training a contrastive learning pretext task with a deep neural network and the SimCLRv2 method.
The second step is fine-tuning this large network with a small amount of labeled data.
The third step is self-training or distillation.
The large pretrained network is used to predict Pseudo-Labels on the complete (unlabeled) data.
These (soft) Pseudo-Labels are then used to train a smaller neural network with CE.
The distillation step could be also performed on the same network as in the pretext task. 
Chen \etal show that even this self-distillation leads to performance improvements\cite{simclrv2}.
\emph{Common ideas: CE, (CE*), CL, PL, PT}

\subsection{One-Stage-Unsupervised}

\subsubsection*{Deep Adaptive Image Clustering (DAC)}
DAC \cite{dac} reformulates unsupervised clustering as a pairwise classification.
Similar to the idea of Pseudo-Labels Chang \etal predict clusters and use these to retrain the network.
The twist is that they calculate the cosine distance between all cluster predictions.
This distance is used to determine whether the input images are similar or dissimilar with a given certainty.
The network is then trained with binary CE on these certain similar and dissimilar input images.
One can interpret these similarities and dissimilarities as Pseudo-Labels for the similarity classification task.
During the training process, they lower the needed certainty to include more images.
As input Chang \etal use a combination of RGB and extracted HOG features.
\emph{Common ideas: PL}

\subsubsection*{Information Maximizing Self-Augmented Training (IMSAT)}
IMSAT \cite{imsat} maximizes MI between the input and output of the model.
As a consistency regularization Hu \etal use CE between an image prediction and an augmented image prediction.
They show that the best augmentation of the prediction can be calculated with VAT \cite{vat}.
The maximization of MI directly on the image input leads to a problem.
For datasets like CIFAR-10, CIFAR-100 \cite{cifar} and STL-10 \cite{stl-10} the color information is too dominant in comparison to the actual content or shape.
As a workaround, Hu \etal use the features generated by a pretrained CNN on ImageNet \cite{imagenet} as input.
\emph{Common ideas: MI, VAT}

\picThree{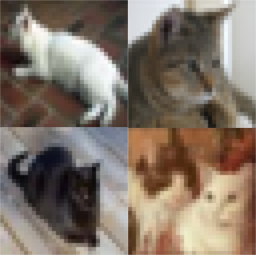}{CIFAR-10}{STL-10}{ILSVRC-2012}{Examples of four random cats in the different datasets to illustrate the difference in quality}

\subsubsection*{Invariant Information Clustering (IIC)}
IIC \cite{iic} is described above as a multi-stage-semi-supervised method.
In comparison to other presented methods, IIC creates usable classifications without fine-tuning the model on labeled data.
The reason for this is that the pretext task is constructed in such a way that label predictions can be extracted directly from the model.
This leads to the conclusion that IIC can also be interpreted as an unsupervised learning method.
\emph{Common ideas: MI, OC}

\subsubsection*{Fuzzy Overclustering (FOC)}
FOC \cite{foc} is described avbove as a multi-stage-semi-supervised method.
Like IIC, FOC can also be seen as an one-stage-unsupervised method because the first stage yields cluster predictions.
\emph{Common ideas: MI, OC}

\subsubsection*{Semantic Clustering by Adopting Nearest Neighbors (SCAN)}
Gansbeke \etal calculate clustering assignments building on self-supervised pretext task by mining the nearest neighbors and using self-labeling.
They propose to use SimCLR \cite{simclr} as a pretext task but show that other pretext tasks \cite{metric-instance,self-rotation} could also be used for this step.
For each sample, the $k$ nearest neighbors are selected in the gained feature space.
The novel semantic clustering loss encourages these samples to be in the same cluster.
Gansbeke \etal noticed that the wrong nearest neighbors have a lower confidence and propose to create Pseudo-Labels on only confident examples for further fine-tuning.
They also show that Overclustering can be successfully used if the number of clusters is not known before.
\emph{Common ideas: OC, PL, PT}

\section{Analysis}
\label{sec:compare}

In this chapter, we will analyze which common ideas are shared or differ between methods.
We will compare the performance of all methods with each other on common deep learning datasets.

\subsection{Datasets}
\label{subsec:datasets}

In this survey, we compare the presented methods on a variety of datasets.
We selected four datasets that were used in multiple papers to allow a fair comparison.
An overview of example images is given in \autoref{fig:datasets}.

\subsubsection*{CIFAR-10 and CIFAR-100} are large datasets of tiny color images with size 32x32 \cite{cifar}.
Both datasets contain 60,000 images belonging to 10 or 100 classes respectively.
The 100 classes in CIFAR-100 can be combined into 20 superclasses.
Both sets provide 50,000 training examples and 10,000 validation examples (image + label).
The presented results are only trained with 4,000 labels for CIFAR-10 and 10,000 labels for CIFAR-100 to represent a semi-supervised case.
If a method uses all labels this is marked independently.

\subsubsection*{STL-10} is dataset designed for unsupervised and semi-supervised learning \cite{stl-10}.
The dataset is inspired by CIFAR-10 \cite{cifar} but provides fewer labels.
It only consists of 5,000 training labels and 8,000 validation labels.
However, 100,000 unlabeled example images are also provided.
These unlabeled examples belong to the training classes and some different classes.
The images are 96x96 color images and were acquired in combination with their labels from ImageNet \cite{imagenet}.

\subsubsection*{ILSVRC-2012} is a subset of ImageNet \cite{imagenet}.
The training set consists of 1.2 million images whereas the validation and the test set include 150,000 images.
These images belong to 1000 object categories.
Due to this large number of categories, it is common to report Top-5 and Top-1 accuracy. 
Top-1 accuracy is the classical accuracy where one prediction is compared to one ground-truth label.
Top-5 accuracy checks if a ground truth label is in a set of at most five predictions.
For further details on accuracy see \autoref{subsec:metric}.
The presented results are only trained with 10\% of labels to represent a semi-supervised case.
If a method uses all labels this is marked independently.

\subsection{Evaluation metrics}
\label{subsec:metric}

We compare the performance of all methods based on their classification score.
This score is defined differently for unsupervised and all other settings.
We follow standard protocol and use the classification accuracy in most cases.
For unsupervised learning, we use cluster accuracy because we need to handle the missing labels during the training.
We need to find the best one-to-one permutations ($\sigma$) from the network cluster predictions to the ground-truth classes.
For $N$ images $x_1, \dots , x_N \in X_l$ with labels $z_{x_i}$ and predictions $f(x_i) \in \mathbb{R}^C$ the accuracy is defined in \autoref{eq:acc} whereas the cluster accuracy is defined in \autoref{eq:cl-acc}.
\begin{equation}
\label{eq:acc}
ACC(x_1, \dots , x_N) = \frac{\sum_{i=1}^{N} \mathds{1}_{z_{x_i} = \argmax_{1 \leq j \leq C}f(x_i)_j}}
{N}
\end{equation}

\begin{equation}
\label{eq:cl-acc}
ACC(x_1, \dots , x_N) = \max_{\sigma} \frac{\sum_{i=1}^{N} \mathds{1}_{z_{x_i} = \sigma (\argmax_{1 \leq j \leq C}f(x_i)_j)}}
{N}
\end{equation}

\subsection{Comparison of methods}
\label{subsec:comparison}

In this subsection, we will compare the methods concerning their used common ideas and performance.
We will summarize the presented results and discuss the underlying trends in the next subsection.

\begin{table*}[tbh]
	\caption{Overview of the methods and their used common ideas ---
		On the left-hand side, the reviewed methods from \autoref{sec:methods} are sorted by the training strategy.
		The top row lists the common ideas.
		Details about the ideas and their abbreviations are given in \autoref{subsec:ideas}.
		The last column and some rows sum up the usage of ideas per method or training strategy.
		\emph{Legend}: 
		(X) The idea is only used indirectly.
		The individual explanations are given in \autoref{sec:methods}.			
	}
	\label{tab:methods}   
	\centering
	
	\resizebox{\textwidth}{!}{%
		
		\begin{tabular}{l c c c c c c c  c c c c c  c}
			\hline\noalign{\smallskip}
			& CE & CE* & EM & CL & KL & MSE & MU & MI & OC  & PT & PL & VAT & \makecell{\emph{Overall}\\ \emph{Sum}}  \\
			\noalign{\smallskip}\hline\noalign{\smallskip}
			\textbf{One-Stage-Semi-Supervised }& & & & & & & & & \\        
			
			\rowcolor{LightGray}
			
			Pseudo-Labels \cite{pseudolabel} &  X & X & & &  &  & &  & & & X & & 3\\
			
			$\pi$ model \cite{temporal-ensembling}  & X& & & &  & X & & & & & & & 2 \\
			
			\rowcolor{LightGray}
			
			Temporal Ensembling \cite{temporal-ensembling} & X & & & &  & X & & & & & & & 2\\			 
			
			Mean Teacher \cite{mean-teacher} & X & & & &  & X & & & & & & & 2\\   
			
			\rowcolor{LightGray} 
			
			VAT \cite{vat} & X & &  & &  &  & & & & & &  X & 2\\
			
			VAT + EntMin \cite{vat} & X & & X & &  &  & &  & & & & X & 3 \\  
			
			\rowcolor{LightGray}      
			
			ICT \cite{ict}  &  X  & & & &  & X & X &  &  & & X & & 4\\        
			
			fast-SWA \cite{fast-SWA} & X & & & &  & X & & & & & & & 2 \\ 			 
			
			\rowcolor{LightGray}

			MixMatch \cite{mixmatch} &  X &  & (X) & & &  X & X  &  & & & X  & & 5\\
			
			EnAET \cite{enaet}  &  X &  & (X) & & X & X & X &  & & AET & X & & 7 \\        
			
			\rowcolor{LightGray}
			
			UDA \cite{uda} & X & & X &  & X & & &  & & &  (X)  & & 4\\
			
			SPamCO \cite{cotraining} & X & X & & & & X & & & & & X & & 4\\
			
			\rowcolor{LightGray}		
			
			ReMixMatch \cite{remixmatch} & X &  X  & (X) &  &  &  &  X & (X)  &  & Rotation & X & & 7\\

			FixMatch \cite{fixmatch}  &  X  & X & (X) & &  &  &  &  & & & X & & 4\\

			\noalign{\smallskip}\hline\noalign{\smallskip}
			\emph{Sum} & 14 & 4 &  6 & 0 & 2 & 8 & 4 & 1 &  0 & 2 & 8 & 2 & 47  \\
			
			\noalign{\smallskip}\hline\noalign{\smallskip}
			\textbf{Multi-Stage-Semi-Supervised}  & & & & & & & & &\\    
			
			\rowcolor{LightGray}
			
			Exemplar \cite{self-exemplar} & X & X & &  & & &  & &  & Augmentation & & & 3\\
			
			Context \cite{self-context} & X & X & &  & & &  & &  & Context & & & 3\\
			
			\rowcolor{LightGray}
			
			Jigsaw \cite{self-jigsaw}  & X  & X & &  & & &  & &  & Jigsaw & & & 3\\ 
			
			DeepCluster \cite{deep-cluster} & X & X & &  & &  &   & & X & Clustering & X & & 5\\
			
			\rowcolor{LightGray}
			
			Rotation \cite{self-rotation}  & X & X & &  & & &  & &  & Rotation & & & 3\\
			
			CPC \cite{cpc,cpcv2} & X & (X) & & X   & & &  & (X) &   &  CL & & & 5 \\

			\rowcolor{LightGray}
			
			CMC \cite{cmc}  & X & (X)  & & X   & & &  & (X) &   &  CL & & & 5\\
			
			DIM \cite{dim} & X & & &  & & & & X &   & MI & & & 3 \\
			
			\rowcolor{LightGray}
			
			AMDIM \cite{amdim} & X & & & & & & & X &   & MI & & & 3\\
			
			DMT \cite{dmt}  & X & X &  &  & & X &  & &  & Metric & X &  & 5 \\
			
			\rowcolor{LightGray}
			
			IIC \cite{iic} & X & & &  & & & & X & X &  MI & & &  4\\
			
			S$^4$L \cite{S4L} & X & X & X &  & &  &  & & & Rotation & X & X & 6\\
			
			\rowcolor{LightGray}	
			
			SimCLR \cite{simclr}  & X & (X) & & & & &  &  &  & CL & & & 3\\				
			
			MoCo \cite{moco}  & X & & &  X & & &  &  &  & Metric & & & 3\\

			\rowcolor{LightGray}	
			
			BYOL \cite{byol}  & X & & &  & & X &  &  &  & Bootstrap & & & 3\\

			FOC \cite{foc}  &  X & (X) & & &  &  &  & X & X & MI &  & & 5 \\  	
			
			\rowcolor{LightGray}		
			
			SimCLRv2 \cite{simclrv2}  & X & (X)  & & X & & &  & & & CL & X & & 5 \\

			\noalign{\smallskip}\hline\noalign{\smallskip}
			\emph{Sum} & 17 & 11 &  1 & 5 & 0 & 1  & 0 & 6 & 3 & 17 & 4 & 1 & 66\\
			
			\noalign{\smallskip}\hline\noalign{\smallskip}
			\textbf{One-Stage-Unsupervised}  & & & & & & & & &\\        
			
			\rowcolor{LightGray}        
			
			DAC \cite {dac} & & & & & &  & & &   & & X & & 1\\
			
			IMSAT \cite{imsat} & & & &  & & & & X  &  & & & X & 2\\
			
			\rowcolor{LightGray}

			IIC \cite{iic} &  & & & & & & & X & X &  MI & & & 3\\

			FOC \cite{foc}  &  &  & & &  &  &  & X & X & MI &  & & 3 \\    
			
			\rowcolor{LightGray}    
			
			SCAN \cite{scan}  &   & & &  &  &  & &  & X & CL & X &  & 3   \\ 
			
			\noalign{\smallskip}\hline\noalign{\smallskip}
			\emph{Sum} & 0 & 0 &  0 & 0 & 0 & 0 & 0 & 3 & 3  & 3 & 2 & 1 & 12 \\
			
			\noalign{\smallskip}\hline\noalign{\smallskip}
			\emph{Overall Sum} & 31 & 54 &  7 & 5 & 2 & 9 & 4 & 10 & 6 & 22 & 14 & 4 & 125 \\
			
			\noalign{\smallskip}\hline
		\end{tabular}%
	}
\end{table*}

\begin{table*}[!tbh]
	\caption{
		Overview of the reported accuracies ---
		The first column states the used method.
		For the supervised baseline, we used the best-reported results which were considered as baselines in the referenced papers.
		The original paper is given in brackets after the score.
		The architecture is given in the second column.
		The last four columns report the Top-1 accuracy score in \% for the respective dataset (See \autoref{subsec:metric} for further details).
		If the results are not reported in the original paper, the reference is given after the result.
		A blank entry represents the fact that no result was reported.
		Be aware that different architectures and frameworks are used which might impact the results. 
		Please see \autoref{subsec:comparison} for a detailed explanation.
		\emph{Legend}:  
		$^\dagger$ 100\% of the labels are used instead of the default value defined in \autoref{subsec:datasets}.
		$^\ddagger$ Multilayer perceptron is used for fine-tuning instead of one fully connected layer.
		Remarks on special architectures and evaluations:
		$^1$ Architecture includes Shake-Shake regularization.
		$^2$ Network uses wider hidden layers.
		$^3$ Method uses ten random classes out of the default 1000 classes.
		$^4$ Network only predicts 20 superclasses instead of the default 100 classes.
		$^5$ Inputs are pretrained ImageNet features.		
		$^6$ Method uses different copies of the network for each input.
		$^7$ The network uses selective kernels \cite{network-kernels}.
	}
	\label{tab:performance}   
	\resizebox{\textwidth}{!}{
		\begin{tabular}{l l c c c c c c c}
			\hline\noalign{\smallskip}
			& Architecture & Publication &  CIFAR-10 & CIFAR-100 & STL-10 & ILSVRC-2012 & ILSVRC-2012 (Top-5)  \\
			\noalign{\smallskip}\hline\noalign{\smallskip}
			Supervised (100\% labels) & Best reported & - &
			98.01\cite{enaet} & 79.82\cite{amdim} & 68.7 \cite{dim} & 85.7 \cite{fixefficientnet} & 97.6 \cite{fixefficientnet} \\
			\noalign{\smallskip}\hline\noalign{\smallskip}

			\textbf{One-Stage-Semi-Supervised}  & & & & & & \\
			
			\rowcolor{LightGray}
			
			Pseudo-Label \cite{pseudolabel} &
			ResNet50v2 \cite{resnet} & 2013 &
			& & &  & 82.41 \cite{S4L} \\
			
			$\pi$ model \cite{temporal-ensembling} &
			CONV-13 & 2017 &
			87.64 & & & & \\
			
			\rowcolor{LightGray}
			
			Temporal Ensembling \cite{temporal-ensembling} &
			CONV-13 & 2017 &
			87.84 & & & & \\
			
			Mean Teacher \cite{mean-teacher} &
			CONV-13 & 2017 & 
			87.69 & & & & \\
			
			\rowcolor{LightGray}

			Mean Teacher \cite{mean-teacher} &
			Wide ResNet-28 & 2017 &
			89.64  & &  & & 90.9\cite{simclrv2} \\

			VAT \cite{vat} & 
			CONV-13 & 2018 &
			88.64 & & & & \\
			
			\rowcolor{LightGray}
			
			VAT \cite{vat} & 
			ResNet50v2 & 2018 & 
			& & & & 82.78 \cite{S4L}\\

			VAT  + EntMin \cite{vat} &
			CONV-13 &  2018 &
			89.45 & & &  &\\
			
			\rowcolor{LightGray}
			
			VAT + EntMin\cite{vat} &
			ResNet50v2 &  2018 &
			86.41 \cite{S4L}  & & & & 83.3 \cite{S4L}   \\

			ICT \cite{ict} & 
			Wide ResNet-28 & 2019 &
			92.34 & & & \\
			
			\rowcolor{LightGray}
			
			ICT \cite{ict} &  
			CONV-13 & 2019 &
			92.71 & & & &\\

			fast-SWA \cite{fast-SWA} &
			CONV-13 & 2019 &
			90.95 & 66.38 & &  & \\

			\rowcolor{LightGray}
			
			fast-SWA \cite{fast-SWA} &
			ResNet-26$^1$ & 2019 &
			93.72 & & & & \\

			MixMatch \cite{mixmatch} &
			Wide ResNet-28  & 2019 &
			95.05 & 74.12 & 94.41 &   \\				
			
			\rowcolor{LightGray}

			EnAET \cite{enaet} &
			Wide ResNet-28 & 2019 &
			94.65 & 73.07 & 95.48 & & \\

			UDA \cite{uda} &
			Wide ResNet-28 & 2019 &
			94.7 & & & 68.66 &  88.52   \\

			\rowcolor{LightGray}

			SPamCo \cite{cotraining} &
			Wide ResNet-28 & 2020 &
			92.95 &  &   & &\\

			ReMixMatch \cite{remixmatch} &
			Wide ResNet-28 & 2020 &
			94.86 & 76.97\cite{fixmatch} &   & &\\
			
			\rowcolor{LightGray}

			FixMatch \cite{fixmatch} &
			Wide ResNet-28 & 2020 &
			95.74 & 77.40 &  &  &    \\

			FixMatch \cite{fixmatch} &
			ResNet-50 & 2020 &
			&  & & 71.46 &  89.13  \\

			\noalign{\smallskip}\hline\noalign{\smallskip}
			\textbf{Multi-Stage-Semi-Supervised }\\
			
			\rowcolor{LightGray}
			
			Exemplar \cite{self-exemplar} &
			ResNet50  & 2015 &
			& & & $46.0^\dagger$ \cite{revisiting-self} & 81.01 \cite{S4L} \\
			
			Context \cite{self-context} &
			ResNet50  &  2015 & 
			& & & $51.4^\dagger$ \cite{revisiting-self}  \\
			
			\rowcolor{LightGray}
			
			Jigsaw \cite{self-jigsaw} &
			AlexNet& 2016 &
			& & & $44.6^\dagger$ \cite{revisiting-self} &  \\
			
			DeepCluster \cite{deep-cluster} &
			AlexNet& 2018 & 
			& & 73.4 \cite{iic} & $41^\dagger$ & \\

			\rowcolor{LightGray}
			
			Rotation \cite{self-rotation} &
			AlexNet & 2018 &
			& & & $55.4^\dagger$ \cite{revisiting-self} &  \\

			Rotation \cite{self-rotation} &
			ResNet50v2 &  2018 & 
			& & &  & 78.53 \cite{S4L} \\

			\rowcolor{LightGray}
			
			CPC \cite{cpcv2} &
			ResNet-170 & 2020 &
			$77.45^\dagger$ \cite{dim} & & $77.81^\dagger$ \cite{dim} & 61.0 & 84.88   \\
			
			CMC \cite{cmc} & 
			AlexNet & 2019 &
			&  & $86.88^ \ddagger$ &   \\
			
			\rowcolor{LightGray}
			
			CMC \cite{cmc} & 
			ResNet-50$^6$  &  2019 &
			&  &  & 70.6  & $89.7^\star$  \\
			
			DIM \cite{dim} &
			AlexNet & 2019 &
			&  & $72.57^ \ddagger$ & &  \\

			\rowcolor{LightGray}
			
			DIM \cite{dim} &
			GAN Discriminator  &  2019 &
			$75.21^ {\dagger\ddagger}$ & $49.74^ {\dagger\ddagger}$ &  &  & \\
			
			AMDIM \cite{amdim} &
			ResNet18  & 2019 &
			$91.3 ^ \dagger$ / $93.6^ {\dagger\ddagger}$ &
			$70.2 ^ \dagger$ / $73.8^ {\dagger\ddagger}$ &
			93.6 / $93.8^ {\ddagger}$ &
			$60.2 ^ \dagger$ / $60.9^ {\dagger\ddagger}$  & \\
			
			\rowcolor{LightGray}
			
			DMT \cite{dmt} &
			Wide ResNet-28 & 2019 &
			88.70  & & & & \\

			IIC \cite{iic} &
			ResNet34 & 2019 &
			& & 85.76 \cite{foc} /  $88.8^\ddagger$ &   \\
			
			\rowcolor{LightGray}

			S$^4$L \cite{S4L} &
			ResNet50v2$^2$  & 2019 &
			& & & 73.21 & $91.23^\star$ \\

			SimCLR \cite{simclr} &
			ResNet50v2$^2$ & 2020 &
			& & & 74.4 \cite{simclrv2} / 76.5$^\dagger$ &  92.6 /  93.2$^{\dagger}$   \\
			
			\rowcolor{LightGray}		
			
			MOCO \cite{moco} &
			ResNet50$^2$ & 2020 &
			& & & 68.6  &    \\ 
			
			MOCO \cite{moco} &
			ResNet50 & 2020 &
			& & & 60.6$^\dagger$ / 71.1$^{\dagger\ddagger}$\cite{mocov2}  &    \\ 
			
			\rowcolor{LightGray}	
			
			BYOL \cite{byol} &
			ResNet200$^2$ & 2020 &
			& & & 77.7  & 93.7   \\

			FOC \cite{foc} &
			ResNet34 & 2020 &
			& & 86.49 &   &    \\  
			
			\rowcolor{LightGray}	   
			
			SimCLRv2 \cite{simclrv2} &
			ResNet-152$^{2,7}$ & 2020 &
			& & &  80.9$^\ddagger$ &  95.5$^\ddagger$  \\

			\noalign{\smallskip}\hline\noalign{\smallskip}
			\textbf{One-Stage-Unsupervised } \\
			
			\rowcolor{LightGray}
			
			DAC \cite{dac} &
			All-ConvNet& 2017 &
			52.18 & 23.75 & 46.99 & 52.72$^3$ &  \\
			
			IMSAT \cite{imsat} &
			Autoencoder$^5$ & 2017 & 
			45.6 & 27.5 & 94.1 & &  \\
			
			\rowcolor{LightGray}
			
			IIC \cite{iic} &
			ResNet34  & 2019 &
			61.7 & $25.7^4$ & 59.6 & & \\   		
			
			FOC \cite{foc} &
			ResNet34 & 2020 &
			& &  60.45  &    \\
			
			\rowcolor{LightGray}
			
			SCAN \cite{scan} &
			ResNet18 & 2020  &
			88.3 & 50.7$^4$ & 80.9 & &  \\

	\end{tabular}}
\end{table*}

\subsubsection*{Comparison concerning used common ideas}
In \autoref{tab:methods} we present all methods and their used common ideas.
Following our definition of common ideas in \autoref{subsec:ideas}, we evaluate only ideas that were used frequently in different papers.
Special details such as the different optimizer for fast-SWA or the used approximation for MI are excluded.
Please see \autoref{sec:methods} for further details.

One might expect that common ideas are used equally between methods and training strategies.
We rather see a tendency that common ideas differ between training strategies.
We will step through all common ideas based on the significance of differentiating the training strategies.

\newpage
\FloatBarrier

A major separation between the training strategies can be based on CE and pretext tasks.
All one-stage-semi-supervised methods use a cross-entropy loss during training whereas only two use additional losses based on pretext tasks.
All multi-stage-semi-supervised methods use a pretext task and use CE for fine-tuning.
All one-stage-semi-supervised methods use no CE and often use a pretext task.
Due to our definition of the training strategies this grouping is expected.

However, further clusters of the common ideas are visible. 
We notice that some common ideas are (almost) solely used by one of the two semi-supervised strategies.
These common ideas are EM, KL, MSE, and MU for one-stage-semi-supervised methods and CL, MI, and OC for multi-stage-semi-supervised methods.
We hypothesize that this shared and different usage of ideas exists due to the different usage of unlabeled data. 
For example, one-stage-semi-supervised methods use the unlabeled and labeled data in the same stage and therefore might need to regularize the training with MSE.

If we compare multi-stage-semi-supervised and one-stage-unsupervised training we notice that MI, OC, and PT are often used in both.
All three of them are not often used with one-stage-semi-supervised training as stated above.
We hypothesize that this similarity arises because most multi-stage-semi-supervised methods have an unsupervised stage followed by a supervised stage.
For the method IIC the authors even proposed to fine-tune the unsupervised method to surpass purely supervised results.
CE*, PL, and VAT are used in several different methods. 
Due to their simple and complementary idea, they can be used in a variety of different methods.
UDA for example uses PL to filter the unlabeled data for useful images.
CE* seems to be more often used by multi-stage-semi-supervised methods.
The parentheses in \autoref{tab:methods}  indicate that they often also motivate another idea like CE$^{-1}$ \cite{foc} or the CL loss \cite{simclr,cpc}.
All in all, we see that the defined training strategies share common ideas inside each strategy and differ in the usage of ideas between them.
We conclude that the definition of the training strategies is not only logical but is also supported by their usage of common ideas.

\subsubsection*{Comparison concerning performance}
We compare the performance of the different methods based on their respective reported results or cross-references in other papers.
For better comparability, we would have liked to recreate every method in a unified setup but this was not feasible.
Whereas using reported values might be the only possible approach, it leads to drawbacks in the analysis.

Kolesnikov \etal showed that changes in the architecture can lead to significant performance boost or drops \cite{revisiting-self}.
They state that 'neither [...] the ranking of architectures [is] consistent across different methods, nor is the ranking of methods consistent across architectures' \cite{revisiting-self}.
Most methods try to achieve comparability with previous ones by a similar setup but over time small differences still aggregate and lead to a variety of used architectures.
Some methods use only early convolutional networks such as AlexNet \cite{imagenet} but others use more modern architectures like Wide ResNet-Architecture \cite{wide-resnet} or Shake-Shake-Regularization \cite{shake-shake}.

Oliver \etal proposed guidelines to ensure more comparable evaluations in semi-supervised learning \cite{realistic-semi-supervised}.
They showed that not following these guidelines may lead to changes in the performance \cite{realistic-semi-supervised}.
Whereas some methods try to follow these guidelines, we cannot guarantee that all methods do so.
This impacts comparability further.
Considering the above-mentioned limitations, we do not focus on small differences but look for general trends and specialties instead.

\autoref{tab:performance} shows the collected results for all presented methods.
We also provide results for the respective supervised baselines reported by the authors.
To keep fair comparability we did not add state-of-the-art baselines with more complex architectures.
\autoref{tab:performance-semi} shows the results for even fewer labels as normally defined in \autoref{subsec:datasets}.

In general, the used architectures become more complex and the accuracies rise over time.
This behavior is expected as new results are often improvements of earlier works.
The changes in architecture may have led to these improvements.
However, many papers include ablation studies and comparisons to only supervised methods to show the impact of their method. 
We believe that a combination of more modern architecture and more advanced methods lead to improvements.

For the CIFAR-10 dataset, almost all multi- or one-stage-semi-supervised methods reach about or over 90\% accuracy.
The best methods MixMatch and FixMatch reach an accuracy of more than 95\% and are roughly three percent worse than the fully supervised baseline.
For the CIFAR-100 dataset, fewer results are reported. 
FixMatch is with about 77\% on this dataset the best method in comparison to the fully supervised baseline of about 80\%.	
Newer methods also provide results for 1000 or even 250 labels instead of 4000 labels.
Especially EnAET, ReMixMatch, and FixMatch stick out since they achieve only 1-2\% worse results with 250 labels instead of with 4000 labels.

For the STL-10 dataset, most methods report a better result than the supervised baseline.
These results are possible due to the unlabeled part of the dataset.
The unlabeled data can only be utilized by semi-, self-, or unsupervised methods.
EnAET achieves the best results with more than 95\%.
FixMatch reports an accuracy of nearly 95\% with only 1000 labels.
This is more than most methods achieve with 5000 labels.

The ILSVRC-2012 dataset is the most difficult dataset based on the reported Top-1 accuracies.
Most methods only achieve a Top-1 accuracy which is roughly 20\% worse than the reported supervised baseline with around 86\%.
Only the methods SimCLR, BYOL, and SimCLRv2 achieve an accuracy that is less than 10\% worse than the baseline.
SimCLRv2 achieves the best accuracy with a Top-1 accuracy of 80.9\% and a Top-5 accuracy of around 96\%.
For fewer labels also SimCLR, BYOL and SimCLRv2 achieve the best results.

\begin{table*}[h]
	\caption{
		Overview of the reported accuracies with fewer labels -
		The first column states the used method.
		The last seven columns report the Top-1 accuracy score in \% for the respective dataset and amount of labels.
		The number is either given as an absolute number or in percent.
		A blank entry represents the fact that no result was reported.
	}
	\label{tab:performance-semi}   
	\resizebox{\textwidth}{!}{
		\begin{tabular}{l c c c c c c c c c}
			\toprule
			& \multicolumn{3}{c}{CIFAR-10}  & \multicolumn{2}{c}{STL-10}   &  \multicolumn{2}{c}{ILSVRC-2012} & \multicolumn{2}{c}{ILSVRC-2012 (Top-5)} \\
			
			\cmidrule(r){2-4} \cmidrule(r){5-6} \cmidrule(r){7-8}  \cmidrule(r){9-10}
			
			& 4000 & 1000 & 250 &  5000  & 1000 & 10\% & 1\% &10\% & 1\%  \\
			
			\midrule
			
			\textbf{One-Stage-Semi-Supervised}  & & & & & & \\
			
			\rowcolor{LightGray}
			
			Mean Teacher \cite{mean-teacher} &
			89.64 & 82.68 & 52.68 & & &  &  & & \\

			ICT \cite{ict} & 
			92.71 & 84.52 & 61.4 \cite{mixmatch} & & & & & &\\
			
			\rowcolor{LightGray}

			MixMatch \cite{mixmatch} &
			
			93.76 & 92.25 &  88.92 &  94.41 &  89.82 & & & & \\
			
			EnAET \cite{enaet} &
			94.65 &  93.05 & 92.4 & 95.48 & 91.96 & & & & \\
			
			\rowcolor{LightGray}
			
			UDA \cite{uda} &
			95.12\cite{fixmatch} &   & 91.18\cite{fixmatch} &  & 92.34\cite{fixmatch} & 68.66 & & 88.52 & \\

			ReMixMatch \cite{remixmatch} &
			94.86 &  94.27 & 93.73  & &  93.82 & & & &\\
			
			\rowcolor{LightGray}
			
			FixMatch \cite{fixmatch} &
			95.74 &   & 94.93  & & 94.83 & 71.46 & & 89.13 &\\

			\noalign{\smallskip}\hline\noalign{\smallskip}
			\textbf{Multi-Stage-Semi-Supervised }\\
			
			\rowcolor{LightGray}

			DMT \cite{dmt} &
			88.70  & & 80.3 & & &  & 58.6 & &\\	
			
			SimCLR \cite{simclr} &			
			& & & & & 74.4\cite{simclrv2} & 63.0\cite{simclrv2} & 92.6 &  85.8 \\

			\rowcolor{LightGray}

			BYOL \cite{byol} &			
			& & & & & 77.7 & 71.2 & 93.7 & 89.5 \\

			SimCLRv2 \cite{simclrv2} &			
			& & & & & 80.9 &  76.6 & 95.5 & 93.4\\

		\end{tabular}
	}
\end{table*}

The unsupervised methods are separated from the supervised baseline by a clear margin of up to 10\%.
SCAN achieves the best results in comparison to the other methods as it builds on the strong pretext task of SimCLR.
This also illustrates the reason for including the unsupervised method in a comparison with semi-supervised methods.
Unsupervised methods do not use labeled examples and therefore are expected to be worse.
However, the data show that the gap of 10\% is not large and that unsupervised methods can benefit from ideas of self-supervised learning.
Some paper report results for even fewer labels as shown in \autoref{tab:performance-semi} which closes the gap to unsupervised learning further.
IMSAT reports an accuracy of about 94\% on STL-10.
Since IMSAT uses pretrained ImageNet features, a superset of STL-10, the results are not directly comparable.

\subsection{Discussion}
\label{subsec:discus}

In this subsection, we discuss the presented results of the previous subsection.
We divide our discussion into three major trends that we identified.
All these trends lead to possible future research opportunities.

\subsubsection*{1. Trend: Real World Applications?}
Previous methods were not scaleable to real-world images and applications and used workarounds e.g. extracted features \cite{imsat} to process real-world images.
Many methods can report a result of over 90\% on CIFAR-10, a simple low-resolution dataset.
Only five methods can achieve a Top-5 accuracy of over 90\% on ILSVRC-2012, a high-resolution dataset.
We conclude that most methods are not scalable to high-resolution and complex image classification problems.
However, the best-reported methods like FixMatch and SimCLRv2 seem to have surpassed the point of only scientific usage and could be applied to real-world classification tasks.

This conclusion applies to real-world image classification tasks with balanced and clearly separated classes.
This conclusion also implicates which real-world issues need to be solved in future research.
Class imbalance \cite{focal-loss,morphocluster} or noisy labels \cite{noisy-data,foc} are not treated by the presented methods.
Datasets with also few unlabeled data points are not considered.
We see that good performance on well-structured datasets does not always transfer completely to real-world datasets \cite{foc}.
We assume that these issues arise due to assumptions that do not hold on real-world datasets like a clear distinction between datapoints \cite{foc} and non-robust hyperparameters like augmentations and batch size \cite{byol}.
Future research has to address these issues so that reduced supervised learning methods can be applied to any real-world datasets.

\subsubsection*{2. Trend: How much supervision is needed?}
We see that the gap between reduced supervised and supervised methods is shrinking.
For CIFAR-10, CIFAR-100 and ILSVRC-2012 we have a gap of less than 5\% left between total supervised and reduced supervised learning.
For STL-10 the reduced supervised methods even surpass the total supervised case by about 20\% due to the additional set of unlabeled data.
We conclude that reduced supervised learning reaches comparable results while using only roughly 10\% of the labels.

In general, we considered a reduction from 100\% to 10\% of all labels.
However, we see that methods like FixMatch and SimCLRv2 achieve comparable results with even fewer labels such as the usage of 1\% of all labels.
For ILSVRC-2012 this is equivalent to about 13 images per class.
FixMatch even achieves a median accuracy of around 65\% for one label per class for the CIFAR-10 dataset\cite{fixmatch}.

The trend that results improve overtime is expected.
But the results indicate that we are near the point where semi-supervised learning needs very few to almost no labels per class (e.g. 10 labels for CIFAR10).
In practice, the labeling cost for unsupervised and semi-supervised will almost be the same for common classification datasets.
Unsupervised methods would need to bridge the performance gap on these classification datasets to be useful anymore.
It is questionable if an unsupervised method can achieve this because it would need to guess what a human wants to have classified even when competing features are available.

We already see that on datasets like ImageNet additional data such as JFT-300M is used to further improve the supervised training \cite{MetaPseudo,BIT,noisy-student}.
These large amounts of data can only be collected without any or weak labels as the collection process has to be automated. 
It will be interesting to investigate if the discussed methods in this survey can also scale to such datasets while using only few labels per class.

We conclude that on datasets with few and a fixed number of classes semi-supervised methods will be more important than unsupervised methods.
However, if we have a lot of classes or new classes should be detected like in few- or zero-shot learning \cite{survey-few,survey-zero,zeroshot,morphocluster} unsupervised methods will still have a lower labeling cost and be of high importance.
This means future research has to investigate how the semi-supervised ideas can be transferred to unsupervised methods as in \cite{iic,scan} and to settings with many, an unknown or rising amount of classes like in \cite{transmatch,MetaPseudo}.

\subsubsection*{3. Trend: Combination of common ideas}
In the comparison, we identified that few common ideas are shared by one-stage-semi-supervised and multi-stage-semi-supervised methods.

We believe there is only a little overlap between these methods due to the different aims of the respective authors.
Many multi-stage-semi-supervised papers focus on creating good representations.
They fine-tune their results only to be comparable.
One-stage-semi-supervised papers aim for the best accuracy scores with as few labels as possible.

If we look at methods like SimCLRv2, EnAET, ReMixMatch, or S$^4$L we see that it can be beneficial to combine different ideas and mindsets.
These methods used a broad range of ideas and also ideas uncommon for their respective training strategy.
$S^4L$ calls their combined approach even "Mix of all models" \cite{S4L} and SimCLRv2 states that "Self-Supervised Methods are Strong Semi-Supervised Learners" \cite{simclrv2}.

We assume that this combination is one reason for their superior performance.
This assumption is supported by the included comparisons in the original papers. 
For example, S$^4$L showed the impact of each method separately as well as the combination of all \cite{S4L}.

Methods like Fixmatch illustrate that it does not need a lot of common ideas to achieve state-of-the-art performance but rather that the selection of the correct ideas and combining them in a meaningful is important.
We identified that some common ideas are not often combined and that the combination of a broad range and unusual ideas can be beneficial.
We believe that the combination of the different common idea is a promising future research field because many reasonable combinations are yet not explored.

\section{Conclusion}

In this paper, we provided an overview of semi-, self-, and unsupervised methods.
We analyzed their difference, similarities, and combinations based on 34 different methods.
This analysis led to the identification of several trends and possible research fields.

We based our analysis on the definition of the different training strategies and common ideas in these strategies.
We showed how the methods work in general, which ideas they use and provide a simple classification.
Despite the difficult comparison of the methods' performances due to different architectures and implementations, we identified three major trends.

Results of over 90\% Top-5 accuracy on ILSVRC-2012 with only 10\% of the labels indicate that semi-supervised methods could be applied to real-world problems.
However, issues like class imbalance and noisy or fuzzy labels are not considered.
More robust methods need to be researched before semi-supervised learning can be applied to real-world issues.

The performance gap between supervised and semi- or self-supervised methods is closing and the number of labels to get comparable results to fully supervised learning is decreasing.
In the future, the unsupervised methods will have almost no labeling cost benefit in comparison to the semi-supervised methods due to these developments.
We conclude that in combination with the fact that semi-supervised methods have the benefit of using labels as guidance unsupervised methods will lose importance.
However, for a large number of classes or an increasing number of classes the ideas of unsupervised are still of high importance and ideas from semi-supervised and self-supervised learning need to be transferred to this setting.

We concluded that one-stage-semi-supervised and multi-stage-semi-supervised training mainly use a different set of common ideas.
Both strategies use a combination of different ideas but there are few overlaps in these techniques.
We identified the trend that a combination of different techniques is beneficial to the overall performance.
In combination with the small overlap between the ideas, we identified possible future research opportunities.

\bibliographystyle{unsrt}
\bibliography{mend}

\newpage
\FloatBarrier

\end{document}